\pdfoutput=1

\documentclass[11pt]{article}

\usepackage[usenames,dvipsnames]{xcolor}
\usepackage[final]{acl}

\usepackage{times}
\usepackage{latexsym}
\usepackage{booktabs}
\usepackage{multirow}
\usepackage{colortbl}
\usepackage{mathtools}
\usepackage{enumitem}
\usepackage{CJKutf8}

\usepackage{listings}
\usepackage[T1]{fontenc}

\usepackage[utf8]{inputenc}

\usepackage{microtype}

\usepackage{inconsolata}

\usepackage{graphicx}
\usepackage{amsmath,amsfonts,bm}

\newcommand{\alignmentmetric}{\mathcal{A}}

\newcommand\blfootnote[1]{%
  \begingroup
  \renewcommand\thefootnote{}\footnote{\hspace{-0.15cm}#1}%
  \addtocounter{footnote}{-1}%
  \endgroup
}

\title{Towards Style Alignment in Cross-Cultural Translation}

\author{Shreya Havaldar$^*$, Adam Stein$^*$, Eric Wong, \& Lyle Ungar \\
  University of Pennsylvania \\
  \texttt{\{shreyah,steinad,exwong,ungar\}@seas.upenn.edu} \\}

\begin{document}
\maketitle
\begin{abstract}
Successful communication depends on the speaker's \textit{intended style} (i.e., what the speaker is trying to convey) aligning with the listener's \textit{interpreted style} (i.e., what the listener perceives). However, cultural differences often lead to misalignment between the two; for example, politeness is often lost in translation. We characterize the ways that LLMs fail to translate style --- biasing translations towards neutrality and performing worse in non-Western languages. We mitigate these failures with RASTA (Retrieval-Augmented STylistic Alignment), a method that leverages learned stylistic concepts to encourage LLM translation to appropriately convey cultural communication norms and align style.\blfootnote{$^*$Equal contribution}\footnote{Code, data, \& prompts available at \url{https://github.com/shreyahavaldar/style_alignment}}

\end{abstract}

\section{Introduction}

\begin{figure}[t]
    \centering
    \includegraphics[width=\columnwidth]{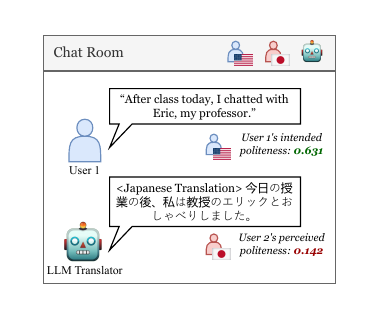}
    \caption{An example of cross-cultural communication facilitated by an LLM. User 1 (American) intends to be polite, but User 2 (Japanese) interprets the message as slightly impolite, given Japanese cultural norms don't typically condone calling a professor by their first name.}
    \label{fig:spirit}
\end{figure}

People from different cultural backgrounds must interact as our world grows more interconnected. 
Machine translation helps promote such intercultural dialogue \cite{shadiev2016facilitating, khasawneh2023potential}, as shown in Figure~\ref{fig:spirit}, with LLMs being increasingly adopted to facilitate translation \citep{Albarino_2024}. These models bridge the \textit{linguistic} gap that may arise during communication; however, another important gap to address is the \textit{cultural} gap \cite{hershcovich-etal-2022-challenges}.

Domains like healthcare and education benefit greatly from shared knowledge \cite{lee2023effectiveness}; however, communication practices differ across cultures \cite{schouten2006cultural, hofstede1986cultural}. These differences introduce unique challenges to cross-cultural communication \cite{moorjani1988semiotic}. For instance, statements that are helpful and appropriate in one culture could be interpreted as critical in another \cite{hall1976beyond}.

Successful communication relies on a speaker's intended style to align with a listener's interpreted style\footnote{Linguistic style reflects the systematic variation in linguistic choices across different contexts and speakers, i.e. features of grammar and vocabulary that signal social identity, attitude, and communicative intent \cite{biber2019register}.} \cite{Thomas1983CrossCulturalPF, tannen1983cross}. However, cultural differences in individuals' thoughts and actions \cite{lehman2004psychology} can lead to a mismatch of such intent and interpretation, so, to successfully bridge the cultural gap, LLMs must translate \textit{style} along with \textit{content} (i.e. the literal meaning).

In this work, we analyze and mitigate translation errors arising from this cultural gap. We focus on style, a key component of communication, and find that modern LLMs often destroy style during translation \cite{kajava2020emotion, troiano2020lost}.

\begin{figure*}[t]
    \centering
    \includegraphics[width=0.9\textwidth]{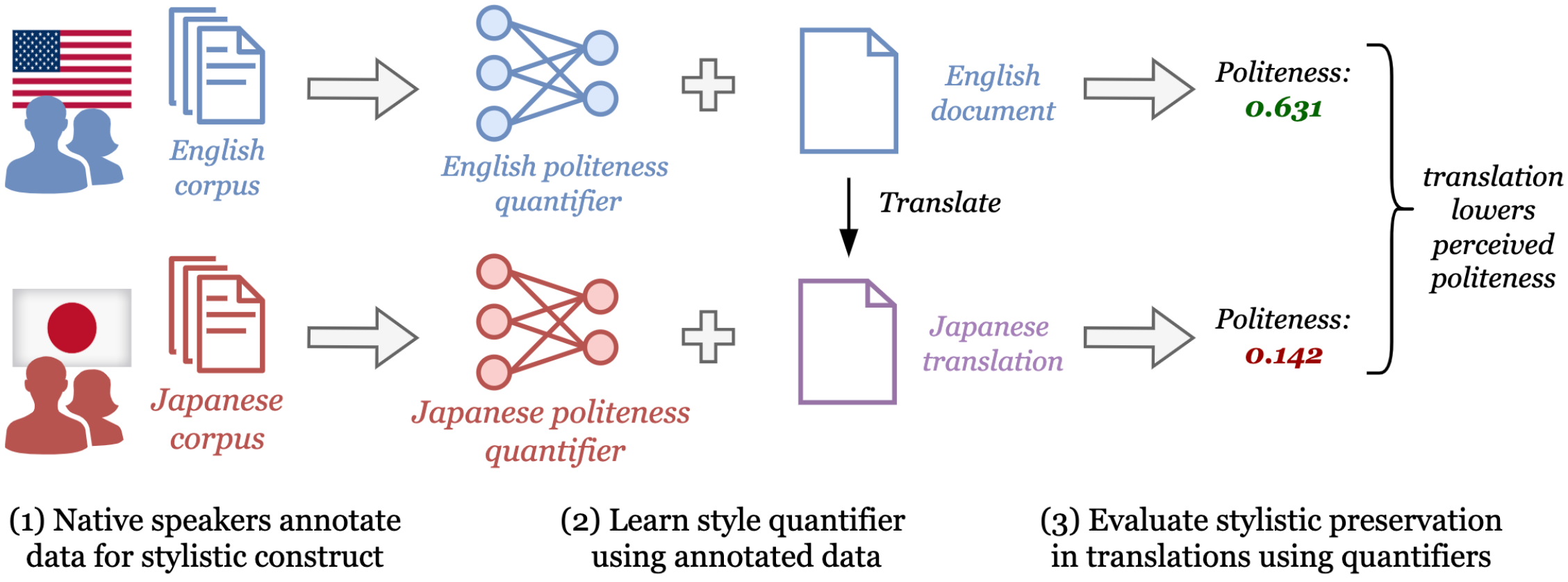}
    \caption{Evaluating how translation affects style. We first select a multilingual style corpus (e.g. the Holistic Politeness Corpus from \citet{havaldar-etal-2023-comparing}) annotated by native speakers from corresponding cultures. We then train separate style quantifiers for each language using text annotated by native speakers. Using these style quantifiers to label the style of the translated text, we can measure how well style is preserved during translation.}
    \label{fig:method}
\end{figure*}

For instance, in Figure~\ref{fig:spirit}, an American (User 1) refers to her professor by his first name. User 1 intends to be polite, and the LLM facilitating the interaction translates the message's content. However, when the Japanese listener (User 2) reads the message, she interprets it as impolite, given that Japanese cultural norms discourage referring to professors by their first names. 

In this work, we characterize failures in style alignment of LLMs. To mitigate these failures, we develop a method where we learn stylistic concepts and leverage available native text to align style during translation without degrading translation quality. Our contributions are as follows:
\begin{itemize}
\vspace{-0.1cm}
    \itemsep=-0.2em
    \item We introduce \textit{style alignment} as a goal for cross-cultural translation. 
    \item Using a variety of datasets and LLMs, we characterize the following style alignment failures:
    \begin{enumerate}
    \vspace{-0.1cm}
        \itemsep=-0.2em
        \item LLMs struggle to translate style, and perform worse in non-Western languages.
        \item LLMs bias translations towards a neutral style, reducing real-world variance.
        \item Current translation metrics poorly reflect style alignment.
    \end{enumerate}
    \item We present RASTA (Retrieval-Augmented STylistic Alignment), a method to \textit{preserve style during translation}, generating translations preferred by bilingual speakers.
\end{itemize}

\section{Designing a Metric for Style Alignment}
\label{sec:evaluation}


In a world where translation models perfectly take into account cultural differences in style, a text's \textit{intended} style in Language $\mathcal{L}_1$ should match the translated text's \textit{interpreted} style in language $\mathcal{L}_2$.

\begin{figure*}[t]
    \centering
    \includegraphics[width=0.9\linewidth]{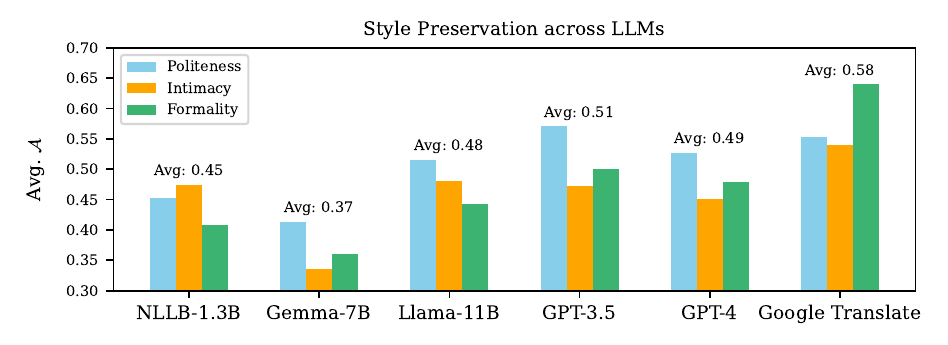}
    \caption{How well do LLMs align style during translation? We show $\alignmentmetric(\mathcal{L}_1, \mathcal{L}_2)$ for today's top LLMs, averaged across language pairs. Google Translate is the best at this task, but all models are far from perfect.}
    \label{fig:style-preservation}
\end{figure*}

\begin{table}[t]
\small
    \centering
    \begin{tabular}{p{5cm}r}
    \toprule
   \textbf{English Examples} & \textbf{Label (0-1)} \\ 
   \midrule
   \multicolumn{2}{l}{\textit{Politeness}} \\
   \midrule
    This has already been debated to death.  & \multirow{1}{*}{0.024}\\ [0.5ex]
    Oh I believe this was already debated! & \multirow{1}{*}{0.465}\\ 
    \midrule
   \multicolumn{2}{l}{\textit{Intimacy}} \\
   \midrule
    Happy New Year! & 0.372\\ [0.5ex]
    Happy new year!!!! Love u & 0.681\\
    \midrule
   \multicolumn{2}{l}{\textit{Formality}} \\
   \midrule
    uh, can i have more details pls? &  0.044\\ [0.5ex]
    Could you provide more details, please? & \multirow{1}{*}{0.989}\\
    \bottomrule
    \end{tabular}
    \caption{Examples of text with the same content, but different style labels from our three style datasets.}
    \label{tab:dataset_examples}
\end{table}

Using this assumption, we quantify the degree of style alignment, $\alignmentmetric$, by measuring the difference between the style of a text in language $\mathcal{L}_1$ and the style of the same text translated to language $\mathcal{L}_2$. For example, we can train two style quantifiers for languages $\mathcal{L}_1$ and $\mathcal{L}_2$, which output a number between $0$ and $1$, measuring the degree of style expressed in an input text. Formally, our style quantifiers are mappings,  $\mathcal{C}_1:\mathcal{L}_1 \rightarrow [0, 1]$ and $\mathcal{C}_2:\mathcal{L}_2 \rightarrow [0, 1]$. Then, given a text $x$, a translator $T: \mathcal{L}_1\rightarrow \mathcal{L}_2$ should ideally satisfy:
\begin{equation}
    \forall x\in \mathcal{L}_1, \mathcal{C}_1(x)=\mathcal{C}_2(T(x)).
\label{eq:equality}
\end{equation}
\noindent Since this strict equality is unlikely to hold in practice, we measure the extent to which this holds by calculating a style alignment score $\alignmentmetric$, i.e. the product-moment correlation between the style of the original and translated text. The style alignment score $\alignmentmetric$ of a corpus $X$ translated from $\mathcal{L}_1$ into $\mathcal{L}_2$ is calculated as follows:
\begin{equation}
   \alignmentmetric(\mathcal{L}_1, \mathcal{L}_2) =  r\left( \mathcal{C}_1\left(X_{\mathcal{L}_1}\right), \mathcal{C}_2\left(T\left(X_{\mathcal{L}_1}\right)\right) \right)
\label{eq:metric}
\end{equation}
\noindent This value measures how well the intended style of a speaker in language $\mathcal{L}_1$ matches the interpretation of a listener in language $\mathcal{L}_2$.

\subsection{Selecting Style Datasets} 

In additional to an alignment metric, we also require data. Specifically, we require style datasets that \textit{span multiple languages}, and are \textit{annotated by native speakers}. 

We select three open source datasets meeting these criteria: 
\begin{enumerate}
\vspace{-0.2cm}
    \itemsep=0em
    \item The Holistic Politeness Dataset \cite{havaldar-etal-2023-comparing}, containing Wikipedia editor interactions annotated for \textbf{politeness} in English, Spanish, Japanese, and Chinese.
    \item Multilingual Tweet Intimacy\footnote{Intimacy refers to whether a text is a communication between strangers vs. people who are emotionally close.} \cite{pei2023semeval}, containing Facebook posts annotated for \textbf{intimacy} in English, Spanish, Brazilian-Portuguese, Italian, French, and Chinese.
    \item GYAFC \cite{rao-tetreault-2018-dear} and XFORMAL \cite{briakou-etal-2021-ola}, containing Yahoo Answers annotated for \textbf{formality} in English, Brazilian-Portuguese, French, and Italian.
\end{enumerate}

Given these datasets contain samples from different languages \textit{in the same domain}, there is minimal difference between the distribution of translated text and native speaker text. Additional statistics for these three datasets are provided in Table~\ref{tab:datasets}.

\subsection{Training Style Quantifiers} 

Automatically measuring intended and interpreted style requires reliable style quantifier models. 

For all of our datasets (politeness, intimacy, and formality), we fine-tune a set of Mistral-7B models \citep{mistral} to output a style score for each sample. We train a separate model per language to encourage understanding of culture-specific stylistic nuances without cross-lingual interference.


The average test RMSE across languages is 0.157, 0.183, and 0.255 for politeness, intimacy, and formality respectively. See Appendix~\ref{app:training} and Table~\ref{tab:rmse} for more details on style quantifier training and evaluation. 



\begin{figure*}[t]
    \centering
    \includegraphics[width=0.65\columnwidth]{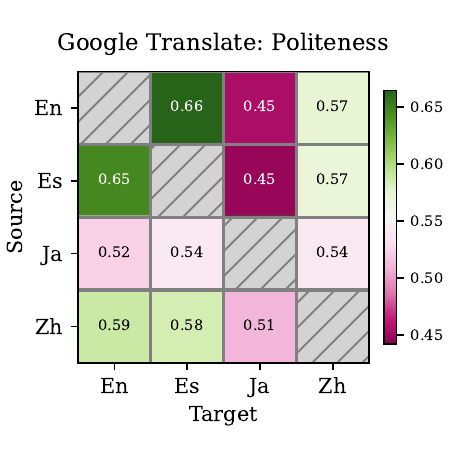}
    \hspace{0.3cm}
    \includegraphics[width=0.65\columnwidth]{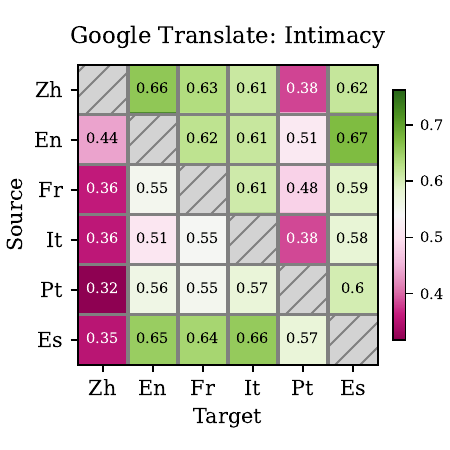}
    \hspace{0.3cm}
     \includegraphics[width=0.65\columnwidth]{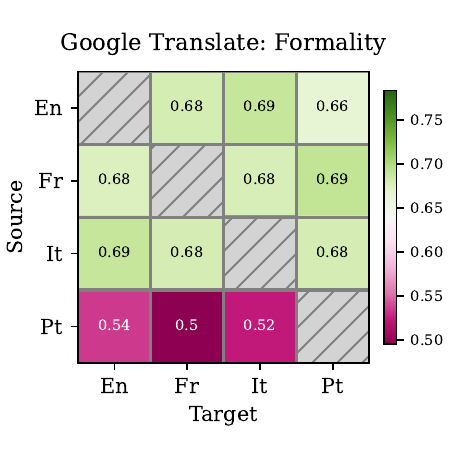}
    \caption{What languages cause Google Translate to struggle the most when aligning style? We show $\alignmentmetric(\mathcal{L}_1, \mathcal{L}_2)$ for each language pair; green indicates above average, and pink indicates below average. Results show style alignment is worst in non-Western languages, raising concerns about successful translation in non-Western cultures.}
    \label{fig:heatmaps}
\end{figure*}

\section{Today's LLMs Struggle to Align Style}
\label{sec:struggle}
With the evaluation framework detailed in Figure~\ref{fig:method}, we can answer the question: 
\textit{How well do state-of-the-art LLMs translate intended style?}

\textbf{Selected models.} \citet{zhu-etal-2024-multilingual} evaluates the translation capabilities of widely-used LLMs. We select the top-performing from this work --- NLLB-1.3B \cite{el2022multilingual}, GPT-3.5, GPT-4, and Google Translate. We also include two high-performing lightweight LLMs --- Gemma-7B \citep{gemmateam2024gemma} and Llama-3.2-11B-Vision-Instruct \citep{llama3}.

\textbf{Evaluation setup.} Each style dataset contains annotated samples from various languages within a single domain. We use the LLM to translate the samples in one language into the other languages in the dataset using a simple prompt: \texttt{Translate the following text from $\mathcal{L}_1 $to $\mathcal{L}_2$: <Text>}, where we provide the source and target languages at runtime. Sampling parameters for all LLMs are provided in Appendix~\ref{app:exp-details}.

Next, for each language pair in the dataset, we calculate the style alignment $\alignmentmetric(\mathcal{L}_1, \mathcal{L}_2)$. Finally, we average $\alignmentmetric(\mathcal{L}_1, \mathcal{L}_2)$ across all language pairs to quantify how well an LLM aligns the given style during translation. Results of this experiment are shown in Figure~\ref{fig:style-preservation}. Google Translate has the highest average $\alignmentmetric(\mathcal{L}_1, \mathcal{L}_2)$ of 0.58 across language pairs and styles. GPT-4, GPT-3.5, and Llama-11B all do comparably well, while the smaller models, NLLB-1.3B and Gemma-7B struggle significantly. These results suggest that style alignment can improve with model size, though even the largest models are far from perfect.

\subsection{Anglocentric Bias in Style Alignment}

Given the suboptimal results in Figure~\ref{fig:style-preservation}, we explore whether the style alignment $\alignmentmetric(\mathcal{L}_1, \mathcal{L}_2)$ is higher for certain language pairs than others. 

We discuss outputs from Google Translate and GPT-4, as the highest-performing and widest used LLMs. Figure~\ref{fig:heatmaps} contains heatmaps showing $\alignmentmetric(\mathcal{L}_1, \mathcal{L}_2)$ for every source language $\mathcal{L}_1$ and target language $\mathcal{L}_2)$. Green indicates a $\alignmentmetric(\mathcal{L}_1, \mathcal{L}_2)$ value above average, while pink indicates below average. GPT-4 heatmaps are shown in Figure~\ref{fig:additional-heatmaps}.

We observe a similar pattern across styles: performance is worst when translating into non-Western languages --- Japanese (ja), Chinese (zh), and Brazilian-Portuguese (pt). These results suggest a mismatch between intended style and interpretation is most likely when communication involves a non-Western language, highlighting issues for LLM translations used in non-Western cultures.

\begin{figure*}[t]
    \centering
    \includegraphics[width=\textwidth]{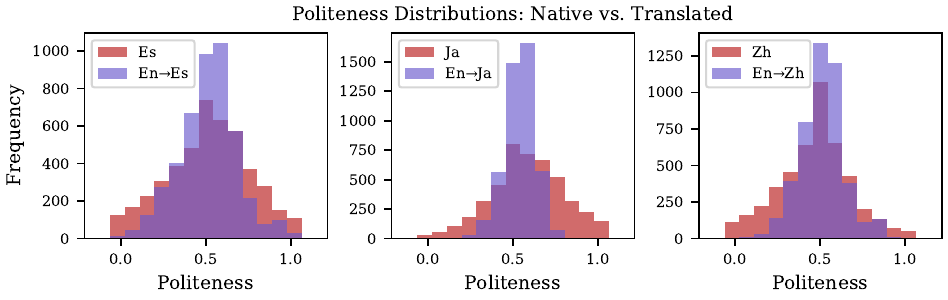}
    \caption{Why do LLMs fail to align style during translation? We plot the politeness distributions of text generated by native speakers vs. translated text. Results show translations skew towards neutral politeness, shrinking standard deviation and reducing the stylistic variance present in the real world.}
    \label{fig:politeness-distribution}
\end{figure*}

\subsection{LLMs Bias Translations towards Neutrality}
Next, we investigate the distribution shift between the style of text written by native speakers of $\mathcal{L}_2$ and the style of text translated to $\mathcal{L}_2$.

We find translations are more ``neutral'' (i.e. having a label of 0.4-0.6) compared to samples written by native speakers. Additionally, stylistic extremes (i.e. 0-0.1 and 0.9-1) exist in text written by native speakers, but rarely occur in translations. 

Figure~\ref{fig:politeness-distribution} plots the distribution of politeness text written by native speakers against English text translated into these languages. We see this neutrality bias in action: the standard deviations of native text (red) are [0.23, 0.20, 0.20] for Spanish, Japanese, and Chinese respectively. However, the translations (blue) have significantly decreased standard deviations of [0.17, 0.09, and 0.13]. 

This phenomenon is particularly problematic for cross-cultural communication involving heightened emotions like enthusiasm or frustration and may lead to confusion or misunderstandings.

\begin{table}[t]
    \centering
    \small
    \begin{tabular}{lrrr}
    \toprule
         &  $\alignmentmetric$ vs. G & $\alignmentmetric$ vs. CK & G vs. CK\\
         \midrule
         Google Trans. & -0.154\phantom{$^*$} & -0.548\phantom{$^*$} & 0.674$^*$\\
         GPT-4 & 0.243\phantom{$^*$} & -0.216\phantom{$^*$} & 0.702$^*$\\
         GPT-3.5 & 0.030\phantom{$^*$} & -0.396$^*$ & 0.648$^*$\\
         Llama 3.2 & 0.070\phantom{$^*$} & -0.171\phantom{$^*$} & 0.788$^*$\\
         NLLB & 0.030\phantom{$^*$} & -0.270$^*$ & 0.889$^*$\\
         Gemma & -0.369$^*$ & -0.181\phantom{$^*$} & 0.287$^*$\\
         \bottomrule
    \end{tabular}
    \caption{Correlations between our style alignment metric $\alignmentmetric$ and traditional translation metrics: GEMBA (G), and \textsc{CometKiwi} (CK). Results shown are the average across all language pairs $\mathcal{L}_1, \mathcal{L}_2$ and models shown in Figure~\ref{fig:style-preservation}. $^*$indicates $p < 0.05$.}
    \label{tab:metric_corr}
\end{table}

\subsection{Modern Metrics Exclude Style Alignment}
Given the numerous ways that today's LLMs fail to translate intended style, this raises the question -- why don't current translation metrics catch this?

In addition to calculating style alignment, we also calculate two popular reference-free translation metrics: GEMBA \cite{kocmi-federmann-2023-large} and \textsc{CometKiwi} \cite{rei2022cometkiwi} on all of our translated data. Finally, we calculate the correlation between these metrics.

Results are shown in Table~\ref{tab:metric_corr}. Across all models, we observe high correlations between the two traditional translation metrics (G and CK), but negative or insignificant correlations when they are compared with $\alignmentmetric$. This indicates traditional metrics poorly evaluate style alignment, providing some insight into why such failures occur.

\begin{figure*}[t]
    \centering
    \includegraphics[width=0.95\textwidth]{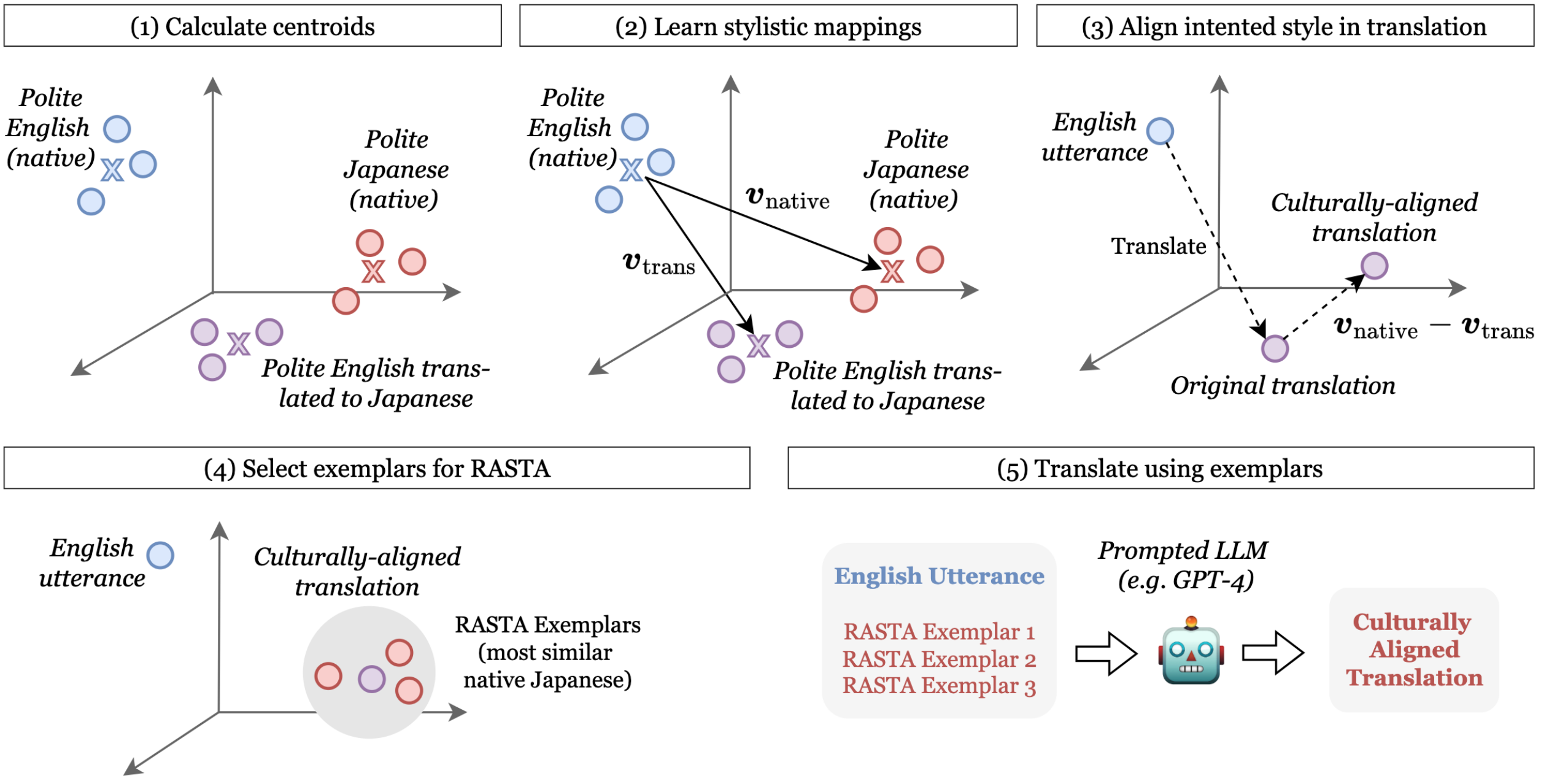}
    \caption{RASTA: our method to align style, shown using English and Japanese. In (1) and (2), we discover stylistic concepts and learn mappings $\bm{v}_{\mathrm{native}}$ from native English $\rightarrow$ native Japanese and $\bm{v}_{\mathrm{trans}}$ from native English → translated Japanese. In (3), we apply an alignment mapping $\bm{v}_{\mathrm{native}} - \bm{v}_{\mathrm{trans}}$ that aligns an input embedding according to cultural communication norms. In (4) and (5), we use the aligned embedding to select the best few-shot exemplars and generate a more culturally aligned translation. \textit{Note that we perform this process separately for every style and language pair, as centroids and exemplars are unique to style level and language.}}
    
    \label{fig:alignment}
\end{figure*}

\section{RASTA: Retrieval-Augmented STylistic Alignment}

We establish in Section~\ref{sec:struggle} that LLMs fail to properly align style during translation. However, past work has found that LLMs succeed at \textit{classifying} style in a multilingual setting, either via fine-tuning or prompting \cite{briakou-etal-2021-ola, plaza2020emoevent, el2022multilingual, havaldar-etal-2023-comparing, srinivasan2022tydip}. 

We explore how this stylistic understanding is encoded in embedding space by probing for stylistic concepts and using them to augment the translation process, developing a method that yields more stylistically-aligned results. 

\subsection{Distinguishing Style in Embedding Space}

To begin, we explore how different styles (e.g. polite vs. rude, formal vs. informal) are encoded differently in embedding space.

To determine if different styles within a language are distinguishable in embedding space, we evaluate if samples are clustered by style.
To calculate embeddings, we use the BGE-M3 model \citep{bge-m3}, a recent high-performing multilingual embedding model. Note that in this section we use known style labels to \textit{analyze} the structure of a given embedding space rather than imposing our own structure.

For a corpus $X_{\mathcal{L}}$ representing text in language $\mathcal{L}$ (e.g. Japanese), let $X_{\mathcal{L},\mathcal{S}}$ be the subset of that corpus labeled with style $\mathcal{S}$ (e.g. Japanese text with a politeness label of ``slightly rude''). Then, the centroid of texts in language $\mathcal{L}$ with style $\mathcal{S}$ is 

\begin{equation}
    \mu(\mathcal{L},\mathcal{S}) = \frac{1}{|X_{\mathcal{L},\mathcal{S}}|}\sum_{x \in X_{\mathcal{L},\mathcal{S}}} \mathcal{E}(x)
\label{eq:centroid}
\end{equation}

\noindent where $\mathcal{E}(x)$ is an embedding function. Subfigure (1) within Figure~\ref{fig:alignment} provides a visualization of this.

\begin{table*}[t]
\small
\renewcommand{\arraystretch}{1.2}
    \centering
    
    \begin{tabular}{@{}lllrrrrrrrrrrrrr@{}}
        \toprule
        \multirow{3}{*}{\textbf{Style}} & \multirow{3}{*}{\textbf{Target}} & \multicolumn{3}{c}{\textbf{Baseline 1: Vanilla}} & \multicolumn{3}{c}{\textbf{Baseline 2: ``Preserve}} & \multicolumn{3}{c}{\multirow{2}{*}{\textbf{RASTA}}} \\
        
        & & \multicolumn{3}{c}{\textbf{Translation}} & \multicolumn{3}{c}{\textbf{Style'' Prompting}} & & \\
        
        \cmidrule(lr){3-5} \cmidrule(lr){6-8} \cmidrule{9-11}
        & & $\alignmentmetric$$\uparrow$ & CK$\uparrow$ & G$\uparrow$ & $\alignmentmetric$$\uparrow$ & CK$\uparrow$ & G$\uparrow$ & $\alignmentmetric$$\uparrow$ & CK$\uparrow$ & G$\uparrow$\\
        \midrule

\multirow{4}{*}{Politeness} & English & 0.61 & 0.80 & 95.33 & 0.60 & 0.80 & 95.39 & 0.70 & 0.78 & 94.61 \\
& Spanish & 0.56 & 0.75 & 95.94 & 0.65 & 0.75 & 96.69 & 0.69 & 0.75 & 96.45 \\
& Japanese & 0.39 & 0.80 & 94.66 & 0.55 & 0.80 & 95.13 & 0.70 & 0.78 & 94.83 \\
& Chinese & 0.55 & 0.76 & 94.80 & 0.61 & 0.76 & 95.03 & 0.70 & 0.75 & 94.64 \\
\cmidrule{2-11}
 & Avg. & 0.53 & 0.78 & 95.18 & 0.60 & 0.78 & 95.56 & 0.70 & 0.77 & 95.13 \\
& RASTA $\Delta$  & \color{Green}{+ 32.1\%} & \color{Red}{-1.3\%} & \color{Green}{+0.0\%} & \color{Green}{+16.7\%} & \color{Red}{-1.3\%} & \color{Red}{-0.4\%} & -- & -- & -- \\ \midrule

\multirow{6}{*}{Intimacy} & English & 0.64 & 0.77 & 93.78 & 0.62 & 0.78 & 94.26 & 0.66 & 0.76 & 93.34 \\
& Spanish & 0.62 & 0.71 & 94.79 & 0.58 & 0.72 & 95.70 & 0.59 & 0.72 & 95.26 \\
& French & 0.38 & 0.71 & 93.86 & 0.57 & 0.72 & 94.95 & 0.60 & 0.72 & 94.48 \\
& Italian & 0.49 & 0.72 & 94.49 & 0.58 & 0.72 & 95.56 & 0.59 & 0.72 & 95.14 \\
& Portuguese & 0.29 & 0.70 & 95.28 & 0.45 & 0.71 & 95.84 & 0.46 & 0.71 & 95.63 \\
& Chinese & 0.28 & 0.70 & 92.24 & 0.37 & 0.71 & 93.43 & 0.39 & 0.71 & 93.05 \\

\cmidrule{2-11}
& Avg. & 0.45 & 0.72 & 94.07 & 0.53 & 0.73 & 94.96 & 0.55 & 0.72 & 94.49 \\
& RASTA $\Delta$  & \color{Green}{+ 22.2\%} & \color{Green}{+0.0\%} & \color{Green}{+0.0\%} & \color{Green}{+1.7\%} & \color{Green}{+1.4\%} & \color{Red}{-0.5\%} & -- & -- & -- \\ \midrule

\multirow{4}{*}{Formality} & En & 0.46 & 0.82 & 97.26 & 0.54 & 0.82 & 97.07 & 0.76 & 0.81 & 96.32 \\
& Fr & 0.44 & 0.81 & 97.11 & 0.66 & 0.81 & 97.35 & 0.75 & 0.80 & 96.98 \\
& It & 0.51 & 0.80 & 97.73 & 0.66 & 0.80 & 98.01 & 0.70 & 0.80 & 97.83 \\
& Pt & 0.50 & 0.79 & 97.75 & 0.72 & 0.80 & 97.97 & 0.78 & 0.78 & 97.36 \\
\cmidrule{2-11}
& Avg. & 0.48 & 0.81 & 97.46 & 0.64 & 0.81 & 97.60 & 0.75 & 0.80 & 97.12 \\
& RASTA $\Delta$  & \color{Green}{+ 56.3\%} & \color{Red}{-1.3\%} & \color{Red}{-0.4\%} & \color{Green}{+17.2\%} & \color{Red}{-1.3\%} & \color{Red}{-0.5\%} & -- & -- & -- \\

\bottomrule
    \end{tabular}
    \caption{GPT-4: Evaluation of RASTA with prompting baselines. We measure the style alignment, $\alignmentmetric$, as well as state-of-the-art reference-free translation quality metrics GEMBA (G) \citep{kocmi-federmann-2023-large} and Comet-Kiwi (CK) \citep{rei2022cometkiwi}. $\alignmentmetric$ is 1 when the interpreted style (i.e. style of translation) exactly matches the intended style (i.e. style of original text) and 0 when there is no alignment whatsoever. For all metrics, a higher score is better.}
    \label{tab:results}
\end{table*}

As seen in Table~\ref{tab:embed_dist}, there is a significant distance between the centroids of \textit{different style} labels within the \textit{same language}, and centroids of the \textit{same style} label in \textit{different languages} (e.g. $\mu(\mathrm{En, polite})\rightarrow\mu(\mathrm{En, rude})$ and $\mu(\mathrm{En, polite})\rightarrow\mu(\mathrm{Ja, polite})$), when compared to the centroids of random subsets of the same size. This distinction implies that LLMs have some encoding of style and how it differs across languages. 

To better understand why LLMs succeed at style classification, but fail to translate style, we also calculate the centroids for translated text. 
Let $\mu( \mathcal{L}_1\rightarrow \mathcal{L}_2, \mathcal{S})$ denote the centroid of text in $\mathcal{L}_2$ with style $\mathcal{S}$ that has been translated from $\mathcal{L}_1$ to $\mathcal{L}_2$. When we embed this translated text, we observe $\mu( \mathcal{L}_1\rightarrow \mathcal{L}_2, \mathcal{S})$ is a significant distance away from $\mu(\mathcal{L}_2, \mathcal{S})$. Table~\ref{tab:embed_dist} suggests a dichotomy between embeddings of text translated into a language $\mathcal{L}_2$ and embeddings of text written by speakers of $\mathcal{L}_2$, despite having the same original style. 

\textbf{Key takeaway.} The distinction between translated and native text in embedding space may be a reason for the subpar results we observe in Figure~\ref{fig:heatmaps}. By learning mappings between these concepts, we can align the embedding of an input text into one that appropriately reflects the the target style.

\subsection{Learning Cultural Alignment Mappings}

Formally, for every source language $\mathcal{L}_1$ and target language $\mathcal{L}_2$, we can calculate the mapping from text in $\mathcal{L}_1$ with style $\mathcal{S}$ to text in $\mathcal{L}_2$ with the same style $\mathcal{S}$. This process is detailed below:
\begin{align*}
    \bm{v}_{\mathrm{native}}(\mathcal{L}_1,\mathcal{L}_2,\mathcal{S}) = \mu(\mathcal{L}_2,\mathcal{S}) - \mu(\mathcal{L}_1,\mathcal{S})
\end{align*} 

\noindent This gives us a mapping in embedding space that encapsulates \textit{linguistic shift}, e.g. moving from polite text in English to polite text in Japanese. 

Similarly, we can calculate the mapping from text in $\mathcal{L}_1$ with style $\mathcal{S}$ to the translations of this text in $\mathcal{L}_2$. This process is detailed below:
\begin{align*}
    \bm{v}_{\mathrm{trans}}(\mathcal{L}_1,\mathcal{L}_2,s) = \mu(\mathcal{L}_1\rightarrow \mathcal{L}_2,s) - \mu(\mathcal{L}_1,s)
\end{align*} 

\noindent This provides an embedding mapping that encapsulates \textit{translation shift}, e.g. moving from polite text in English to its translation in Japanese. Subfigure (2) in Figure~\ref{fig:alignment} shows examples of such mappings.

After calculating $\bm{v}_{\mathrm{native}}$ and $\bm{v}_{\mathrm{trans}}$, we can calculate exactly how the embedding of a translation in $\mathcal{L}_2$ would need to be transformed to exist in the same part of embedding space as native text in $\mathcal{L}_2$.

In theory, this transformation would eliminate the cultural gap between text in $\mathcal{L}_1$ and its translation in $\mathcal{L}_2$, as Equation~\ref{eq:equality} would hold. In other words, the intended style in $\mathcal{L}_1$ would exactly match the interpreted style $\mathcal{L}_2$. This cultural alignment mapping, the key component of our RASTA algorithm, can be calculated as follows:
\begin{align*}
    \bm{v}_{\mathrm{align}}: = \bm{v}_{\mathrm{native}}(\mathcal{L}_1,\mathcal{L}_2,\mathcal{S}) - \bm{v}_{\mathrm{trans}}(\mathcal{L}_1,\mathcal{L}_2,\mathcal{S}) 
\end{align*} 

\noindent Subfigure (3) within Figure~\ref{fig:alignment} details this alignment.

\subsection{Preserving Intended Style in Translations}

Using this cultural alignment mapping, we can embed and transform any piece of text in source language $\mathcal{L}_1$ to exactly where it \textit{should} lie in embedding space in order for Equation~\ref{eq:equality} to hold. 

\textbf{Exemplar selection.} From there, we find the five exemplars in the training set of $\mathcal{L}_2$ with the highest cosine similarity to this transformed embedding. We use these exemplars as few-shot examples during translation time, encouraging the LLM to analyze how native speakers of $\mathcal{L}_2$ express the style $\mathcal{S}$ of the original text in $\mathcal{L}_1$. The few-shot generation prompt is provided in Appendix~\ref{app:prompts}.

Subfigures (4) and (5) within Figure~\ref{fig:alignment} show these final steps of RASTA.

\section{Experiments \& Results}

Using our translation framework, RASTA, we re-translate the test sets of all three style datasets. We then evaluate the effectiveness of RASTA by comparing our translations against a set of baselines and measuring both style alignment and traditional translation quality. Since RASTA requires prompting access, we focus on GPT-4 and Llama.

\textbf{Baseline 1.}
As described in Section~\ref{sec:struggle}, our first baseline uses a \textit{vanilla, straightforward prompt}, providing the LLM with instructions to translate a given text into the target language.

\textbf{Baseline 2.}
To evaluate the need for RASTA over sophisticated prompting techniques, we also write a thorough prompt detailing how word-for-word translation may lead to stylistic misalignment, and \textit{explicitly providing instructions to preserve the given style}. The full prompt is in Appendix~\ref{app:prompts}.

\paragraph{Evaluation metrics.} For RASTA and our two baselines, we calculate $\alignmentmetric$ to measure style alignment and calculate GEMBA and \textsc{CometKiwi} to measure traditional translation quality.

\subsection{RASTA Improves Style Alignment}

Table~\ref{tab:results} provides results for RASTA using GPT-4. We observe a significant increase in style alignment using RASTA (up to 56\% improvement), without significant degradation in GEMBA or \textsc{CometKiwi} (under 1.5\% degradation). We show examples of improved translations in Table~\ref{tab:rasta-examples}.

The solid increase in performance from Baseline 2 also indicates that \textit{RASTA better aligns style than what can be done by using a well-written prompt.} We also provide results for Llama-3.2-11B in Table~\ref{tab:llama3.2}, and observe similar patterns of improvement over baselines.

\begin{figure}[t]
    \centering
    \includegraphics[width=0.49\columnwidth]{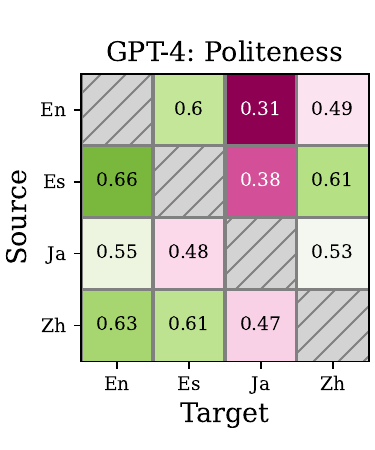}
    \includegraphics[width=0.49\columnwidth]{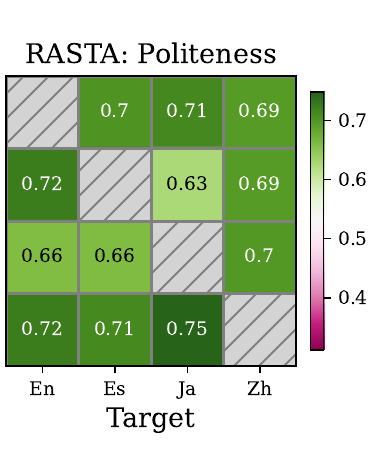}
    \caption{Comparing style alignment between vanilla GPT-4 and RASTA using GPT-4. RASTA improves performance when translating into non-Western languages and reduces the performance gap from 0.35 to 0.12.}
    \label{fig:rasta-heatmap-politeness}
\end{figure}

\paragraph{Mitigating Anglocentric bias.} As shown in Figure~\ref{fig:rasta-heatmap-politeness}, RASTA drastically improves politeness alignment for Chinese and Japanese translation. We observe similar improvement in non-Western languages for intimacy and formality as well (see Figures~\ref{fig:additional-heatmaps} and~\ref{fig:rasta-heatmaps}), highlighting RASTA's ability to de-bias translation performance.

\paragraph{Preserving native speaker variance.} RASTA also decreases the tendency of translations to be neutral. For GPT-4, the standard deviation of the translated politeness distribution (see Figure~\ref{fig:politeness-distribution}) increases from vanilla prompting to RASTA as follows: [0.14, 0.10, 0.10] $\rightarrow$ [0.18, 0.13, 0.15] for Spanish, Japanese, Chinese, a 36\% average increase and a much better reflection of the politeness variance occurring in native text.

\subsection{Humans Prefer RASTA Translations} To confirm whether native speakers view RASTA translations as an improvement, we run an annotation study on Prolific. 

\paragraph{Study setup.} For every language pair \{English, $\mathcal{L}$\} in our politeness and formality datasets\footnote{We do not include intimacy, as a large portion of the dataset contains sexually explicit content.}, we recruit a set of 3 bilingual annotators on Prolific, for a total of 18 annotators across languages and datasets. Annotators are required to have selected $\mathcal{L}$ as their first language and be fully fluent in English. Annotators are then shown 30 randomly selected samples, along with their RASTA translations and their ``preserve style'' prompting translations, and asked to select which translation better preserves style. 

\paragraph{Annotation results.} Annotators select RASTA a majority of the time -- 61\% of the time for politeness, and 63\% of the time for formality, indicating RASTA aligns with native speaker intuition. See Appendix~\ref{app:annotation} for study details, annotator agreement, and additional results.

Overall, \textit{RASTA succeeds in improving style alignment while still outputting translations that fully preserve content and meaning.} 

\section{Related Work}

\textbf{Controlling style in generations: } Past style transfer work controls for style via sampling and ranking \cite{niu-etal-2017-study} and fine-tuning \cite{rippeth2022controlling, garcia2021towards}. Similar approaches have been used to control style in translation \cite{niu2018multi, schioppa-etal-2021-controlling, sennrich2016controlling, nadejde-etal-2022-cocoa}. RASTA differs from these works as style is not forced on the output; rather, it matches that of the input. TextSETTR \cite{riley2021textsettr} employs a similar problem setup but only focuses on English, thus eliminating the need for culture-specific style understanding. Low-resource benchmarks \cite{krishna2022fewshotcontrollablestyletransfer, mukherjee2024multilingualtextstyletransfer} employ language-specific techniques, but therefore lack the generalizability of RASTA.


\textbf{Culturally-aware translation:} Prior work in this space involves benchmark datasets that account for cultural differences in names/objects \cite{yao2024benchmarkingmachinetranslationcultural, peskov-etal-2021-adapting-entities}, geographic locations \cite{riley2023frmt}, social norms \cite{huang-yang-2023-culturally}, dialogue \cite{li2024cultureparkboostingcrossculturalunderstanding}, along with investigations of translating time \cite{shwartz-2022-good}, recipes \cite{cao2024cultural}, and idioms \cite{li2024translate}. However, the techniques described in these works focus on \textit{entity replacement} via fine-tuning or post-processing, thus modifying content. Conversely, RASTA focuses on modifying \textit{style} to increase cultural awareness, keeping content intact. 

\textbf{Retrieval-augmented translation:}
Recent work uses in-context learning and retrieval-augmented generation (RAG) for LLM translation \citep{bulte2019neural, xu2020boosting, agrawal-etal-2023-context, garcia2023unreasonable}.
Beyond translation pairs, retrieval from external knowledge bases \citep{conia-etal-2024-towards, chen2024crat}, unstructured data \citep{wang2024retrieval}, and search engines \citep{gu2018search} have been used to improve translation. RAG has also been combined with multi-step prompting \citep{he-etal-2024-exploring}.
Unlike these approaches, \textit{we do not require translation pairs,} instead used a learned alignment mapping to directly retrieve examples in the target language.


\section{Conclusion}
We introduce style alignment as a goal for translation, given that success in communication requires an alignment of intended and interpreted style. We characterize the failures of today's LLMs in aligning style: poor performance, Anglocentric bias, skewing towards neutrality, and low correlation with standard translation quality. To mitigate these failures, we introduce RASTA, a method that leverages stylistic concepts and in tandem with native text to generate culturally-aligned translations.

Our work provides essential insights and methodologies to enhance LLMs' capabilities in cross-cultural communication, establishing a foundation for future research in this field.

\section{Limitations}

In order to calculate our ``ground truth'' style labels for our alignment metric $\alignmentmetric$, we train models (regressors), which are imperfect. The average RMSE of our style quantifiers (across language pairs and styles) is 0.195. Though this suggests high overall performance, certain predictions may be incorrect. This work would benefit from another annotation study including native speaker annotations of our style quantifier models. Unfortunately, due to resource constraints, multiple annotation studies were not feasible for this work.

We only look at three styles in this work: politeness, intimacy, and formality. Future work is needed to assess whether RASTA improves style alignment for the many other styles that exist. Additionally, content and style are deeply intertwined; it may not always be possible to transform style without also modifying content, making perfect style alignment unachievable. 

As with any retrieval-augmented generation method, bias and error in our embedding model may propagate up and result in suboptimal few-shot exemplars. Additionally, since we use cosine similarity to select exemplars, we choose examples close in both content and style, as the embedding encodes both of these constructs. This may provide a suboptimal level of variance in our exemplars.

Though the authors spend time prompt-tuning, we only run experiments with a single prompt for RASTA and each of our baselines; final translations are prompt-dependent and may vary with different prompts. Future work is needed to establish the effect of prompt wording and the number of few-shot exemplars on the final RASTA translations.

This work would be strengthened by explicit inclusion of social and cultural norms to modify translations. However, there are many open questions when determining how to use them for translation: how to distinguish their influence on style vs. content, how to extract norms from style corpora, how to determine which norms to provide to the model, etc. This inclusion is out of scope for this paper and we leave it for future work.

On a higher level, style is subjective, even within languages and cultures, so the ground truth style label likely differs from person to person. This work treats the style of a given text as a static value, which abstracts away all real-world subjectivity.

\section{Ethical Considerations}

Our definition of ``culturally-aware'' translation hinges on style alignment; however, culture is deeply complex and consists of many more communication patterns/norms beyond style. While aligning style is a step in the right direction, we acknowledge that it is an incomplete step to full cultural alignment of LLM generations.

We also only study high-resource languages in this work, as we are limited to what languages are available in open-source style datasets. Future work is needed to determine the effectiveness of RASTA on low-resource languages.

Additionally, given we use an LLM to generate the final translations, inherent bias in or fairness concerns associated with the LLM may propagate up into our generated RASTA translations.

\bibliography{custom}

\begin{thebibliography}{58}
\providecommand{\natexlab}[1]{#1}

\bibitem[{Agrawal et~al.(2023)Agrawal, Zhou, Lewis, Zettlemoyer, and Ghazvininejad}]{agrawal-etal-2023-context}
Sweta Agrawal, Chunting Zhou, Mike Lewis, Luke Zettlemoyer, and Marjan Ghazvininejad. 2023.
\newblock \href {https://doi.org/10.18653/v1/2023.findings-acl.564} {In-context examples selection for machine translation}.
\newblock In \emph{Findings of the Association for Computational Linguistics: ACL 2023}, pages 8857--8873, Toronto, Canada. Association for Computational Linguistics.

\bibitem[{Albarino(2024)}]{Albarino_2024}
Seyma Albarino. 2024.
\newblock \href {https://slator.com/translation-companies-accelerating-adoption-large-language-models-alc-survey/} {Translation companies accelerating adoption of large language models, alc survey}.

\bibitem[{Biber and Conrad(2009)}]{biber2019register}
Douglas Biber and Susan Conrad. 2009.
\newblock \emph{Register, genre, and style}.
\newblock Cambridge University Press.

\bibitem[{Briakou et~al.(2021)Briakou, Lu, Zhang, and Tetreault}]{briakou-etal-2021-ola}
Eleftheria Briakou, Di~Lu, Ke~Zhang, and Joel Tetreault. 2021.
\newblock \href {https://doi.org/10.18653/v1/2021.naacl-main.256} {Ol{\'a}, bonjour, salve! {XFORMAL}: A benchmark for multilingual formality style transfer}.
\newblock In \emph{Proceedings of the 2021 Conference of the North American Chapter of the Association for Computational Linguistics: Human Language Technologies}, pages 3199--3216, Online. Association for Computational Linguistics.

\bibitem[{Bulte and Tezcan(2019)}]{bulte2019neural}
Bram Bulte and Arda Tezcan. 2019.
\newblock Neural fuzzy repair: Integrating fuzzy matches into neural machine translation.
\newblock In \emph{57th Annual Meeting of the Association-for-Computational-Linguistics (ACL)}, pages 1800--1809.

\bibitem[{Cao et~al.(2024)Cao, Kementchedjhieva, Cui, Karamolegkou, Zhou, Dare, Donatelli, and Hershcovich}]{cao2024cultural}
Yong Cao, Yova Kementchedjhieva, Ruixiang Cui, Antonia Karamolegkou, Li~Zhou, Megan Dare, Lucia Donatelli, and Daniel Hershcovich. 2024.
\newblock Cultural adaptation of recipes.
\newblock \emph{Transactions of the Association for Computational Linguistics}, 12:80--99.

\bibitem[{Chen et~al.(2024{\natexlab{a}})Chen, Xiao, Zhang, Luo, Lian, and Liu}]{bge-m3}
Jianlv Chen, Shitao Xiao, Peitian Zhang, Kun Luo, Defu Lian, and Zheng Liu. 2024{\natexlab{a}}.
\newblock \href {https://arxiv.org/abs/2402.03216} {Bge m3-embedding: Multi-lingual, multi-functionality, multi-granularity text embeddings through self-knowledge distillation}.
\newblock \emph{Preprint}, arXiv:2402.03216.

\bibitem[{Chen et~al.(2024{\natexlab{b}})Chen, Meng, Zhang, Zhang, and Zhou}]{chen2024crat}
Meiqi Chen, Fandong Meng, Yingxue Zhang, Yan Zhang, and Jie Zhou. 2024{\natexlab{b}}.
\newblock Crat: A multi-agent framework for causality-enhanced reflective and retrieval-augmented translation with large language models.
\newblock \emph{arXiv preprint arXiv:2410.21067}.

\bibitem[{Conia et~al.(2024)Conia, Lee, Li, Minhas, Potdar, and Li}]{conia-etal-2024-towards}
Simone Conia, Daniel Lee, Min Li, Umar~Farooq Minhas, Saloni Potdar, and Yunyao Li. 2024.
\newblock \href {https://doi.org/10.18653/v1/2024.emnlp-main.914} {Towards cross-cultural machine translation with retrieval-augmented generation from multilingual knowledge graphs}.
\newblock In \emph{Proceedings of the 2024 Conference on Empirical Methods in Natural Language Processing}, pages 16343--16360, Miami, Florida, USA. Association for Computational Linguistics.

\bibitem[{Conneau et~al.(2019)Conneau, Khandelwal, Goyal, Chaudhary, Wenzek, Guzm{\'{a}}n, Grave, Ott, Zettlemoyer, and Stoyanov}]{xlmroberta}
Alexis Conneau, Kartikay Khandelwal, Naman Goyal, Vishrav Chaudhary, Guillaume Wenzek, Francisco Guzm{\'{a}}n, Edouard Grave, Myle Ott, Luke Zettlemoyer, and Veselin Stoyanov. 2019.
\newblock \href {https://arxiv.org/abs/1911.02116} {Unsupervised cross-lingual representation learning at scale}.
\newblock \emph{CoRR}, abs/1911.02116.

\bibitem[{Dettmers et~al.(2024)Dettmers, Pagnoni, Holtzman, and Zettlemoyer}]{qlora}
Tim Dettmers, Artidoro Pagnoni, Ari Holtzman, and Luke Zettlemoyer. 2024.
\newblock Qlora: Efficient finetuning of quantized llms.
\newblock \emph{Advances in Neural Information Processing Systems}, 36.

\bibitem[{Dubey et~al.(2024)Dubey, Jauhri, Pandey, Kadian, Al-Dahle, Letman, Mathur, Schelten, Yang, Fan et~al.}]{llama3}
Abhimanyu Dubey, Abhinav Jauhri, Abhinav Pandey, Abhishek Kadian, Ahmad Al-Dahle, Aiesha Letman, Akhil Mathur, Alan Schelten, Amy Yang, Angela Fan, et~al. 2024.
\newblock The llama 3 herd of models.
\newblock \emph{arXiv preprint arXiv:2407.21783}.

\bibitem[{El-Alami et~al.(2022)El-Alami, El~Alaoui, and Nahnahi}]{el2022multilingual}
Fatima-zahra El-Alami, Said~Ouatik El~Alaoui, and Noureddine~En Nahnahi. 2022.
\newblock A multilingual offensive language detection method based on transfer learning from transformer fine-tuning model.
\newblock \emph{Journal of King Saud University-Computer and Information Sciences}, 34(8):6048--6056.

\bibitem[{Garcia et~al.(2023)Garcia, Bansal, Cherry, Foster, Krikun, Johnson, and Firat}]{garcia2023unreasonable}
Xavier Garcia, Yamini Bansal, Colin Cherry, George Foster, Maxim Krikun, Melvin Johnson, and Orhan Firat. 2023.
\newblock The unreasonable effectiveness of few-shot learning for machine translation.
\newblock In \emph{International Conference on Machine Learning}, pages 10867--10878. PMLR.

\bibitem[{Garcia et~al.(2021)Garcia, Constant, Guo, and Firat}]{garcia2021towards}
Xavier Garcia, Noah Constant, Mandy Guo, and Orhan Firat. 2021.
\newblock Towards universality in multilingual text rewriting.
\newblock \emph{arXiv preprint arXiv:2107.14749}.

\bibitem[{Gemma-Team(2024)}]{gemmateam2024gemma}
Gemma-Team. 2024.
\newblock \href {https://arxiv.org/abs/2403.08295} {Gemma: Open models based on gemini research and technology}.
\newblock \emph{Preprint}, arXiv:2403.08295.

\bibitem[{Gu et~al.(2018)Gu, Wang, Cho, and Li}]{gu2018search}
Jiatao Gu, Yong Wang, Kyunghyun Cho, and Victor~OK Li. 2018.
\newblock Search engine guided neural machine translation.
\newblock In \emph{Proceedings of the AAAI Conference on Artificial Intelligence}, volume~32.

\bibitem[{Hall(1976)}]{hall1976beyond}
Edward~T Hall. 1976.
\newblock \emph{Beyond culture}.
\newblock Anchor.

\bibitem[{Havaldar et~al.(2023)Havaldar, Pressimone, Wong, and Ungar}]{havaldar-etal-2023-comparing}
Shreya Havaldar, Matthew Pressimone, Eric Wong, and Lyle Ungar. 2023.
\newblock \href {https://doi.org/10.18653/v1/2023.emnlp-main.419} {Comparing styles across languages}.
\newblock In \emph{Proceedings of the 2023 Conference on Empirical Methods in Natural Language Processing}, pages 6775--6791, Singapore. Association for Computational Linguistics.

\bibitem[{He et~al.(2024)He, Liang, Jiao, Zhang, Yang, Wang, Tu, Shi, and Wang}]{he-etal-2024-exploring}
Zhiwei He, Tian Liang, Wenxiang Jiao, Zhuosheng Zhang, Yujiu Yang, Rui Wang, Zhaopeng Tu, Shuming Shi, and Xing Wang. 2024.
\newblock \href {https://doi.org/10.1162/tacl_a_00642} {Exploring human-like translation strategy with large language models}.
\newblock \emph{Transactions of the Association for Computational Linguistics}, 12:229--246.

\bibitem[{Hershcovich et~al.(2022)Hershcovich, Frank, Lent, de~Lhoneux, Abdou, Brandl, Bugliarello, Cabello~Piqueras, Chalkidis, Cui, Fierro, Margatina, Rust, and S{\o}gaard}]{hershcovich-etal-2022-challenges}
Daniel Hershcovich, Stella Frank, Heather Lent, Miryam de~Lhoneux, Mostafa Abdou, Stephanie Brandl, Emanuele Bugliarello, Laura Cabello~Piqueras, Ilias Chalkidis, Ruixiang Cui, Constanza Fierro, Katerina Margatina, Phillip Rust, and Anders S{\o}gaard. 2022.
\newblock \href {https://doi.org/10.18653/v1/2022.acl-long.482} {Challenges and strategies in cross-cultural {NLP}}.
\newblock In \emph{Proceedings of the 60th Annual Meeting of the Association for Computational Linguistics (Volume 1: Long Papers)}, pages 6997--7013, Dublin, Ireland. Association for Computational Linguistics.

\bibitem[{Hofstede(1986)}]{hofstede1986cultural}
Geert Hofstede. 1986.
\newblock Cultural differences in teaching and learning.
\newblock \emph{International Journal of intercultural relations}, 10(3):301--320.

\bibitem[{Huang and Yang(2023)}]{huang-yang-2023-culturally}
Jing Huang and Diyi Yang. 2023.
\newblock \href {https://doi.org/10.18653/v1/2023.findings-emnlp.509} {Culturally aware natural language inference}.
\newblock In \emph{Findings of the Association for Computational Linguistics: EMNLP 2023}, pages 7591--7609, Singapore. Association for Computational Linguistics.

\bibitem[{Jiang et~al.(2023)Jiang, Sablayrolles, Mensch, Bamford, Chaplot, Casas, Bressand, Lengyel, Lample, Saulnier et~al.}]{mistral}
Albert~Q Jiang, Alexandre Sablayrolles, Arthur Mensch, Chris Bamford, Devendra~Singh Chaplot, Diego de~las Casas, Florian Bressand, Gianna Lengyel, Guillaume Lample, Lucile Saulnier, et~al. 2023.
\newblock Mistral 7b.
\newblock \emph{arXiv preprint arXiv:2310.06825}.

\bibitem[{Kajava et~al.(2020)Kajava, {\"O}hman, Hui, and Tiedemann}]{kajava2020emotion}
Kaisla Kajava, Emily {\"O}hman, Piao Hui, and J{\"o}rg Tiedemann. 2020.
\newblock Emotion preservation in translation: Evaluating datasets for annotation projection.
\newblock In \emph{Digital Humanities in the Nordic Countries}, pages 38--50. CEUR.

\bibitem[{Khasawneh(2023)}]{khasawneh2023potential}
Mohamad Ahmad~Saleem Khasawneh. 2023.
\newblock The potential of ai in facilitating cross-cultural communication through translation.
\newblock \emph{Journal of Namibian Studies: History Politics Culture}, 37:107--130.

\bibitem[{Kocmi and Federmann(2023)}]{kocmi-federmann-2023-large}
Tom Kocmi and Christian Federmann. 2023.
\newblock \href {https://aclanthology.org/2023.eamt-1.19} {Large language models are state-of-the-art evaluators of translation quality}.
\newblock In \emph{Proceedings of the 24th Annual Conference of the European Association for Machine Translation}, pages 193--203, Tampere, Finland. European Association for Machine Translation.

\bibitem[{Krishna et~al.(2022)Krishna, Nathani, Garcia, Samanta, and Talukdar}]{krishna2022fewshotcontrollablestyletransfer}
Kalpesh Krishna, Deepak Nathani, Xavier Garcia, Bidisha Samanta, and Partha Talukdar. 2022.
\newblock \href {https://arxiv.org/abs/2110.07385} {Few-shot controllable style transfer for low-resource multilingual settings}.
\newblock \emph{Preprint}, arXiv:2110.07385.

\bibitem[{Lee(2023)}]{lee2023effectiveness}
Sangmin-Michelle Lee. 2023.
\newblock The effectiveness of machine translation in foreign language education: a systematic review and meta-analysis.
\newblock \emph{Computer Assisted Language Learning}, 36(1-2):103--125.

\bibitem[{Lehman et~al.(2004)Lehman, Chiu, and Schaller}]{lehman2004psychology}
Darrin~R Lehman, Chi-yue Chiu, and Mark Schaller. 2004.
\newblock Psychology and culture.
\newblock \emph{Annu. Rev. Psychol.}, 55:689--714.

\bibitem[{Li et~al.(2024{\natexlab{a}})Li, Teney, Yang, Wen, Xie, and Wang}]{li2024cultureparkboostingcrossculturalunderstanding}
Cheng Li, Damien Teney, Linyi Yang, Qingsong Wen, Xing Xie, and Jindong Wang. 2024{\natexlab{a}}.
\newblock \href {https://arxiv.org/abs/2405.15145} {Culturepark: Boosting cross-cultural understanding in large language models}.
\newblock \emph{Preprint}, arXiv:2405.15145.

\bibitem[{Li et~al.(2024{\natexlab{b}})Li, Chen, Yuan, Wu, Yang, Tao, and Xiao}]{li2024translate}
Shuang Li, Jiangjie Chen, Siyu Yuan, Xinyi Wu, Hao Yang, Shimin Tao, and Yanghua Xiao. 2024{\natexlab{b}}.
\newblock Translate meanings, not just words: Idiomkb’s role in optimizing idiomatic translation with language models.
\newblock In \emph{Proceedings of the AAAI Conference on Artificial Intelligence}, volume~38, pages 18554--18563.

\bibitem[{Moorjani and Field(1988)}]{moorjani1988semiotic}
Angela Moorjani and Thomas~T Field. 1988.
\newblock Semiotic and sociolinguistic paths to understanding culture.
\newblock In \emph{Toward a new integration of language and culture. Reports of the Northeast Conference on the Teaching of Foreign Languages}, pages 25--45.

\bibitem[{Mukherjee et~al.(2024)Mukherjee, Ojha, Bansal, Alok, McCrae, and Dušek}]{mukherjee2024multilingualtextstyletransfer}
Sourabrata Mukherjee, Atul~Kr. Ojha, Akanksha Bansal, Deepak Alok, John~P. McCrae, and Ondřej Dušek. 2024.
\newblock \href {https://arxiv.org/abs/2405.20805} {Multilingual text style transfer: Datasets \& models for indian languages}.
\newblock \emph{Preprint}, arXiv:2405.20805.

\bibitem[{Nadejde et~al.(2022)Nadejde, Currey, Hsu, Niu, Federico, and Dinu}]{nadejde-etal-2022-cocoa}
Maria Nadejde, Anna Currey, Benjamin Hsu, Xing Niu, Marcello Federico, and Georgiana Dinu. 2022.
\newblock \href {https://doi.org/10.18653/v1/2022.findings-naacl.47} {{C}o{C}o{A}-{MT}: A dataset and benchmark for contrastive controlled {MT} with application to formality}.
\newblock In \emph{Findings of the Association for Computational Linguistics: NAACL 2022}, pages 616--632, Seattle, United States. Association for Computational Linguistics.

\bibitem[{Niu et~al.(2017)Niu, Martindale, and Carpuat}]{niu-etal-2017-study}
Xing Niu, Marianna Martindale, and Marine Carpuat. 2017.
\newblock \href {https://doi.org/10.18653/v1/D17-1299} {A study of style in machine translation: Controlling the formality of machine translation output}.
\newblock In \emph{Proceedings of the 2017 Conference on Empirical Methods in Natural Language Processing}, pages 2814--2819, Copenhagen, Denmark. Association for Computational Linguistics.

\bibitem[{Niu et~al.(2018)Niu, Rao, and Carpuat}]{niu2018multi}
Xing Niu, Sudha Rao, and Marine Carpuat. 2018.
\newblock Multi-task neural models for translating between styles within and across languages.
\newblock \emph{arXiv preprint arXiv:1806.04357}.

\bibitem[{Pei et~al.(2023)Pei, Silva, Bos, Liu, Neves, Jurgens, and Barbieri}]{pei2023semeval}
Jiaxin Pei, V{\'\i}tor Silva, Maarten Bos, Yozen Liu, Leonardo Neves, David Jurgens, and Francesco Barbieri. 2023.
\newblock Semeval-2023 task 9: Multilingual tweet intimacy analysis.
\newblock In \emph{Proceedings of the 17th International Workshop on Semantic Evaluation (SemEval-2023)}, pages 2235--2246.

\bibitem[{Peskov et~al.(2021)Peskov, Hangya, Boyd-Graber, and Fraser}]{peskov-etal-2021-adapting-entities}
Denis Peskov, Viktor Hangya, Jordan Boyd-Graber, and Alexander Fraser. 2021.
\newblock \href {https://doi.org/10.18653/v1/2021.findings-emnlp.315} {Adapting entities across languages and cultures}.
\newblock In \emph{Findings of the Association for Computational Linguistics: EMNLP 2021}, pages 3725--3750, Punta Cana, Dominican Republic. Association for Computational Linguistics.

\bibitem[{Plaza-del Arco et~al.(2020)Plaza-del Arco, Strapparava, Lopez, and Mart{\'\i}n-Valdivia}]{plaza2020emoevent}
Flor~Miriam Plaza-del Arco, Carlo Strapparava, L~Alfonso~Urena Lopez, and M~Teresa Mart{\'\i}n-Valdivia. 2020.
\newblock Emoevent: A multilingual emotion corpus based on different events.
\newblock In \emph{Proceedings of the 12th Language Resources and Evaluation Conference}, pages 1492--1498.

\bibitem[{Rao and Tetreault(2018)}]{rao-tetreault-2018-dear}
Sudha Rao and Joel Tetreault. 2018.
\newblock \href {https://doi.org/10.18653/v1/N18-1012} {Dear sir or madam, may {I} introduce the {GYAFC} dataset: Corpus, benchmarks and metrics for formality style transfer}.
\newblock In \emph{Proceedings of the 2018 Conference of the North {A}merican Chapter of the Association for Computational Linguistics: Human Language Technologies, Volume 1 (Long Papers)}, pages 129--140, New Orleans, Louisiana. Association for Computational Linguistics.

\bibitem[{Rei et~al.(2022)Rei, Treviso, Guerreiro, Zerva, Farinha, Maroti, De~Souza, Glushkova, Alves, Lavie et~al.}]{rei2022cometkiwi}
Ricardo Rei, Marcos Treviso, Nuno~M Guerreiro, Chrysoula Zerva, Ana~C Farinha, Christine Maroti, Jos{\'e}~GC De~Souza, Taisiya Glushkova, Duarte~M Alves, Alon Lavie, et~al. 2022.
\newblock Cometkiwi: Ist-unbabel 2022 submission for the quality estimation shared task.
\newblock \emph{arXiv preprint arXiv:2209.06243}.

\bibitem[{Riley et~al.(2021)Riley, Constant, Guo, Kumar, Uthus, and Parekh}]{riley2021textsettr}
Parker Riley, Noah Constant, Mandy Guo, Girish Kumar, David~C Uthus, and Zarana Parekh. 2021.
\newblock Textsettr: Few-shot text style extraction and tunable targeted restyling.
\newblock In \emph{Proceedings of the 59th Annual Meeting of the Association for Computational Linguistics and the 11th International Joint Conference on Natural Language Processing (Volume 1: Long Papers)}, pages 3786--3800.

\bibitem[{Riley et~al.(2023)Riley, Dozat, Botha, Garcia, Garrette, Riesa, Firat, and Constant}]{riley2023frmt}
Parker Riley, Timothy Dozat, Jan~A Botha, Xavier Garcia, Dan Garrette, Jason Riesa, Orhan Firat, and Noah Constant. 2023.
\newblock Frmt: A benchmark for few-shot region-aware machine translation.
\newblock \emph{Transactions of the Association for Computational Linguistics}, 11:671--685.

\bibitem[{Rippeth et~al.(2022)Rippeth, Agrawal, and Carpuat}]{rippeth2022controlling}
Elijah Rippeth, Sweta Agrawal, and Marine Carpuat. 2022.
\newblock Controlling translation formality using pre-trained multilingual language models.
\newblock \emph{arXiv preprint arXiv:2205.06644}.

\bibitem[{Schioppa et~al.(2021)Schioppa, Vilar, Sokolov, and Filippova}]{schioppa-etal-2021-controlling}
Andrea Schioppa, David Vilar, Artem Sokolov, and Katja Filippova. 2021.
\newblock \href {https://doi.org/10.18653/v1/2021.emnlp-main.535} {Controlling machine translation for multiple attributes with additive interventions}.
\newblock In \emph{Proceedings of the 2021 Conference on Empirical Methods in Natural Language Processing}, pages 6676--6696, Online and Punta Cana, Dominican Republic. Association for Computational Linguistics.

\bibitem[{Schouten and Meeuwesen(2006)}]{schouten2006cultural}
Barbara~C Schouten and Ludwien Meeuwesen. 2006.
\newblock Cultural differences in medical communication: a review of the literature.
\newblock \emph{Patient education and counseling}, 64(1-3):21--34.

\bibitem[{Sennrich et~al.(2016)Sennrich, Haddow, and Birch-Mayne}]{sennrich2016controlling}
Rico Sennrich, Barry Haddow, and Alexandra Birch-Mayne. 2016.
\newblock Controlling politeness in neural machine translation via side constraints.
\newblock In \emph{15th Annual Conference of the North American Chapter of the Association for Computational Linguistics: Human Language Technologies}, pages 35--40. Association for Computational Linguistics.

\bibitem[{Shadiev and Huang(2016)}]{shadiev2016facilitating}
Rustam Shadiev and Yueh-Min Huang. 2016.
\newblock Facilitating cross-cultural understanding with learning activities supported by speech-to-text recognition and computer-aided translation.
\newblock \emph{Computers \& Education}, 98:130--141.

\bibitem[{Shwartz(2022)}]{shwartz-2022-good}
Vered Shwartz. 2022.
\newblock \href {https://doi.org/10.18653/v1/2022.findings-acl.224} {Good night at 4 pm?! time expressions in different cultures}.
\newblock In \emph{Findings of the Association for Computational Linguistics: ACL 2022}, pages 2842--2853, Dublin, Ireland. Association for Computational Linguistics.

\bibitem[{Srinivasan and Choi(2022)}]{srinivasan2022tydip}
Anirudh Srinivasan and Eunsol Choi. 2022.
\newblock Tydip: A dataset for politeness classification in nine typologically diverse languages.
\newblock \emph{arXiv preprint arXiv:2211.16496}.

\bibitem[{Tannen(1983)}]{tannen1983cross}
Deborah Tannen. 1983.
\newblock \emph{Cross-Cultural Communication.}
\newblock ERIC.

\bibitem[{Thomas(1983)}]{Thomas1983CrossCulturalPF}
Jenny~A. Thomas. 1983.
\newblock \href {https://api.semanticscholar.org/CorpusID:35667508} {Cross-cultural pragmatic failure}.
\newblock \emph{Applied Linguistics}, 4:91--112.

\bibitem[{Troiano et~al.(2020)Troiano, Klinger, and Pad{\'o}}]{troiano2020lost}
Enrica Troiano, Roman Klinger, and Sebastian Pad{\'o}. 2020.
\newblock Lost in back-translation: Emotion preservation in neural machine translation.
\newblock In \emph{Proceedings of the 28th International Conference on Computational Linguistics}, pages 4340--4354.

\bibitem[{Wang et~al.(2024)Wang, Meng, Zhang, and Zhou}]{wang2024retrieval}
Jiaan Wang, Fandong Meng, Yingxue Zhang, and Jie Zhou. 2024.
\newblock Retrieval-augmented machine translation with unstructured knowledge.
\newblock \emph{arXiv preprint arXiv:2412.04342}.

\bibitem[{Xu et~al.(2020)Xu, Crego, and Senellart}]{xu2020boosting}
Jitao Xu, Josep-Maria Crego, and Jean Senellart. 2020.
\newblock Boosting neural machine translation with similar translations.
\newblock In \emph{Annual Meeting of the Association for Computational Linguistics}, pages 1570--1579. Association for Computational Linguistics.

\bibitem[{Yao et~al.(2024)Yao, Jiang, Bobinac, Yang, and Hu}]{yao2024benchmarkingmachinetranslationcultural}
Binwei Yao, Ming Jiang, Tara Bobinac, Diyi Yang, and Junjie Hu. 2024.
\newblock \href {https://arxiv.org/abs/2305.14328} {Benchmarking machine translation with cultural awareness}.
\newblock \emph{Preprint}, arXiv:2305.14328.

\bibitem[{Zhu et~al.(2024)Zhu, Liu, Dong, Xu, Huang, Kong, Chen, and Li}]{zhu-etal-2024-multilingual}
Wenhao Zhu, Hongyi Liu, Qingxiu Dong, Jingjing Xu, Shujian Huang, Lingpeng Kong, Jiajun Chen, and Lei Li. 2024.
\newblock \href {https://doi.org/10.18653/v1/2024.findings-naacl.176} {Multilingual machine translation with large language models: Empirical results and analysis}.
\newblock In \emph{Findings of the Association for Computational Linguistics: NAACL 2024}, pages 2765--2781, Mexico City, Mexico. Association for Computational Linguistics.

\end{thebibliography}

\clearpage
\appendix

\renewcommand{\thefigure}{A\arabic{figure}}
\setcounter{figure}{0}
\renewcommand{\thetable}{A\arabic{table}} 
\setcounter{table}{0}

\section{Style Quantifiers: Training Details}
\label{app:training}

To train the style quantifiers, we finetune Mistral-7B \citep{mistral} on the training datasets for each language in each available style. Since the style labels are normalized values between zero and one, these are regression models rather than classifiers.

\citet{havaldar-etal-2023-comparing}  find that multilingual vs. monolingual LLMs focus on different features when learning to classify style. Taking this into account, we fine-tune a unique model for each language, as we do not want cross-lingual interference.

Since we train a separate model for each language and style, we train a total of 14 models. Due to resource constraints, we could not finetune the entire Mistral 7B model, so we use QLoRA \citep{qlora} to finetune the model on two NVIDIA-A100 GPUs. Specifically, we use 4 bit quantization and a LoRA dimension of 4. We used MSE loss, a learning rate of 5e-5, a batch size of 8, and trained for 5 epochs. The test RMSE of our models is reported in Table~\ref{tab:rmse}. We additionally experimented with finetuning the XLM-RoBERTa-Base model \citep{xlmroberta}, but we found that the Mistral-7B model resulted in lower RMSEs for almost all styles and languages.

\section{LLM Sampling Parameters}
\label{app:exp-details}

For generation from Llama-3.2 and Gemma, we used a temperature of 0.6 and top-$p$ of 0.9. For generation from GPT-3.5 and GPT-4, we used a temperature of 1.0 and top-$p$ of 1.0. We keep all default sampling parameters to evaluate LLMs in their commonly used form.

\begin{table}[t]
    \centering
    \small   
    \begin{tabular}{lr}
    \toprule
        \textbf{Computation} &  \textbf{Distance}\\
        \midrule
        \multicolumn{2}{l}{Different styles within the same language} \\
        \midrule
         $\left| \mu(\mathcal{L}, \mathrm{polite}) - \mu(\mathcal{L}, \mathrm{rude}) \right|$ & 2.01$\pm$0.26\\
         \midrule
        \multicolumn{2}{l}{Identical style across different languages} \\
        \midrule
         $\left| \mu(\mathcal{L}_1, \mathrm{polite}) - \mu(\mathcal{L}_2, \mathrm{polite}) \right|$ & 2.91$\pm$0.34\\
         $\left| \mu(\mathcal{L}_1, \mathrm{rude}) - \mu(\mathcal{L}_2, \mathrm{rude}) \right|$ & 2.77$\pm$0.35\\
         \midrule
        \multicolumn{2}{l}{Translated vs. native speaker generated within} \\
        \multicolumn{2}{l}{the same style and language} \\
        \midrule
         $\left| \mu(\mathcal{L}_1 \rightarrow \mathcal{L}_2, \mathrm{polite}) - \mu(\mathcal{L}_2, \mathrm{polite}) \right|$ & 2.66$\pm$0.25\\
         $\left| \mu(\mathcal{L}_1 \rightarrow \mathcal{L}_2, \mathrm{rude}) - \mu(\mathcal{L}_2, \mathrm{rude}) \right|$ & 2.32$\pm$0.20\\
         \midrule
        \multicolumn{2}{l}{Baseline: random subsets of identical size} \\
        \midrule
         $\left| X_{\text{rand}} - X_{\text{rand}} \right|$ & 0.52$\pm$0.02\\
    \bottomrule
    \end{tabular}
    \caption{Distances in embedding space of politeness dataset subsets. Polite corresponds to the top 20\% of politeness scores and rude is the bottom 20\% of scores.}
    \label{tab:embed_dist}
\end{table}

\begin{table}[ht]
    \centering
    \small
    \begin{tabular}{llr}
    \toprule
         \textbf{Style} & \textbf{Language} & \textbf{Test RMSE} \\
         \midrule
         & English & 0.155\\
         Politeness & Spanish & 0.164\\
         \textit{(0-1 scale)} & Japanese & 0.160\\
         & Chinese & 0.147\\
         \midrule
         & English & 0.127\\
         & Spanish & 0.156\\
         Intimacy & French & 0.179\\
         \textit{(0-1 scale)} & Italian & 0.226\\
         & Portuguese & 0.244\\
         & Chinese & 0.156\\
         \midrule
         & English & 0.262\\
         Formality & French & 0.202\\
         \textit{(0,1 binary)} & Italian & 0.289\\
         & Portuguese & 0.268\\
        \bottomrule
    \end{tabular}
    \caption{Test RMSE for our style quantifiers, showing significantly better-than-random performance.}
    \label{tab:rmse}
\end{table}

\begin{table*}[t]
    \centering
    \small
    \begin{tabular}{p{4.5cm}rrl}
    \toprule
        \textbf{Dataset} & \textbf{Train Size} & \textbf{Test Size} & \textbf{Contained Languages} \\
        \midrule
        Politeness \cite{havaldar-etal-2023-comparing} & 18238 & 2280 & English, Spanish, Japanese, Chinese\\
        Intimacy \cite{pei2023semeval} & 9491 & 2370 & English, Spanish, Portuguese, Italian, French, Chinese\\       
        Formality \cite{rao-tetreault-2018-dear, briakou-etal-2021-ola} & 6392 & 1598 & English, French, Italian, Portuguese\\
    \bottomrule
    \end{tabular}
    \caption{Dataset statistics. For the three datasets used in this work, we show the number of train and test samples and the included languages.}
    \label{tab:datasets}
\end{table*}

\begin{table*}[t]
    \centering
    \small
    \begin{tabular}{lllp{9.2cm}r}
    \toprule 
            \multicolumn{5}{c}{\color{Red}{\textsc{\textbf{WARNING: the examples in this table contain explicit and potentially offensive content.}}}} \\[1ex]
    \toprule
         \textbf{Style} & \textbf{Method} & \textbf{Lang.} & \textbf{Sentence} & \textbf{Label (0-1)}\\
         \midrule
         \multirow{13}{*}{Politeness} & \multirow{4}{*}{Source} & \multirow{4}{*}{Ja} & \begin{CJK}{UTF8}{min}
なるほど、ここまでの経緯とは別の態度のご発言ということなのでしょうか？ コメント依頼を出したのは私ではないから、外すことを求められても困ります。 図の問題は、私のこだわりではありません。
\end{CJK} & \multirow{4}{*}{0.25} \\ [9ex]

& \multirow{4}{*}{Vanilla} & \multirow{4}{*}{En} & I see, so you’re saying this comment shows a different attitude from the course of events so far? I’m not the one who requested the comment, so I’m in trouble if I’m asked to remove it. The problem with the diagram is not my concern. & \multirow{4}{*}{0.566} \\ [9ex]

& \multirow{5}{*}{RASTA} & \multirow{5}{*}{En} & So, are you saying that this statement reflects a different attitude than what's happened up to now? It is rather inconvenient since I didn't ask for the comment to be made, and it's not up to me to decide to eliminate it. The issue with the diagram, it's important to note, is not my personal fixation. & \multirow{5}{*}{0.277}\\ [10ex]

\midrule
\multirow{3}{*}{Intimacy} & Source & Es & Con el pitó bien duro $\#$heteros http & 0.84 \\ [1ex]

& Vanilla & En & With the whistle blowing hard $\#$heteros http & 0.40\\ [1ex]

& RASTA & En & With a rock hard cock $\#$straight http & 0.867\\ [0.5ex]

\midrule
\multirow{6}{*}{Formality} & \multirow{2}{*}{Source} & \multirow{2}{*}{Pt} & Não tenha a ilusão de que ele vai se separar pra ficar com você (são raríssimos esses casos). & \multirow{2}{*}{0.0}\\ [3.5ex]

& \multirow{2}{*}{Vanilla} & \multirow{2}{*}{En} & Don’t be under the illusion that he is going to leave to be with you (such cases are extremely rare). & \multirow{2}{*}{1.0}\\ [3.5ex]

& \multirow{2}{*}{RASTA} & \multirow{2}{*}{En} & Don't kid yourself that he's gonna leave her for you (those cases are super rare). & \multirow{2}{*}{0.11} \\ [1ex]
\bottomrule
    \end{tabular}
    \caption{Qualitative examples comparing RASTA to the vanilla translation baseline with GPT-4. For all examples, RASTA results in a higher level of style alignment than the vanilla translation.}
    \label{tab:rasta-examples}
\end{table*}

\begin{figure*}[t]
    \centering
    \includegraphics[width=0.65\columnwidth]{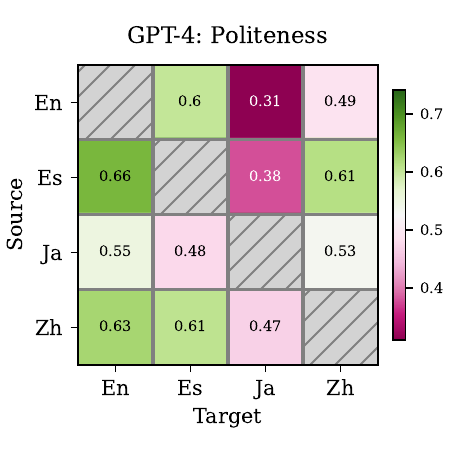}
    \hspace{0.3cm}
    \includegraphics[width=0.65\columnwidth]{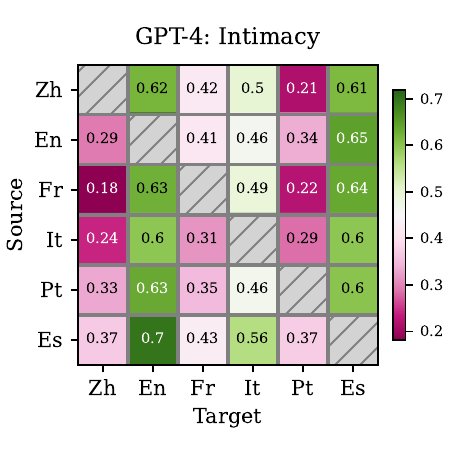}
    \hspace{0.3cm}
     \includegraphics[width=0.65\columnwidth]{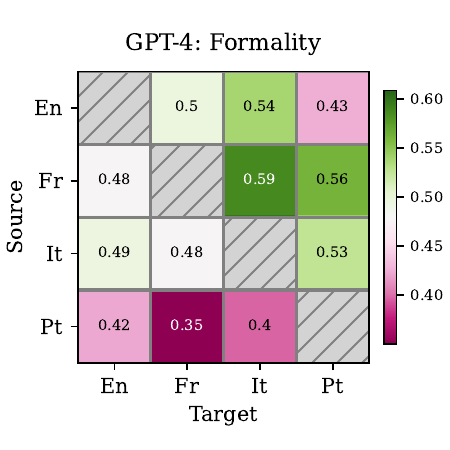}
    \caption{Heatmaps for GPT-4. We show $\alignmentmetric(\mathcal{L}_1, \mathcal{L}_2)$ for each language pair; green indicates above average, and pink indicates below average. Results show style alignment is worst in non-Western languages, raising concerns about successful translation in non-Western cultures.}
    \label{fig:additional-heatmaps}
\end{figure*}

\begin{figure*}[t]
    \centering
    \includegraphics[width=0.65\columnwidth]{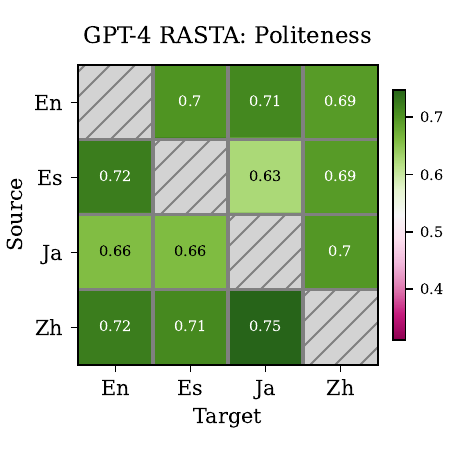}
    \hspace{0.3cm}
    \includegraphics[width=0.65\columnwidth]{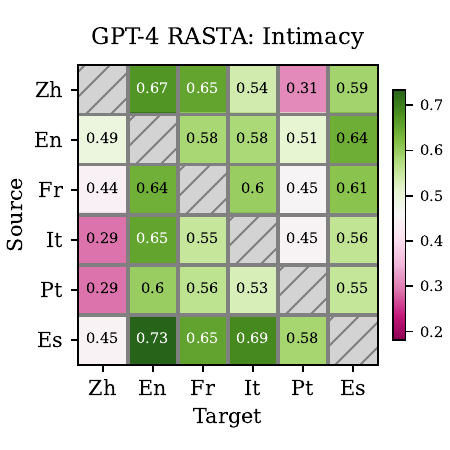}
    \hspace{0.3cm}
     \includegraphics[width=0.65\columnwidth]{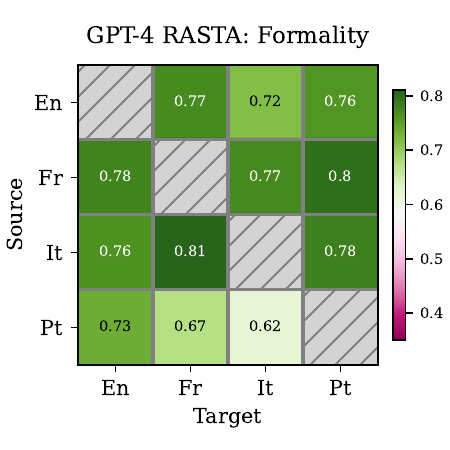}
    \caption{Heatmaps for RASTA with GPT-4. We show $\alignmentmetric(\mathcal{L}_1, \mathcal{L}_2)$ for each language pair; green indicates above average, and pink indicates below average. Results show style alignment is worst in non-Western languages, raising concerns about successful translation in non-Western cultures.}
    \label{fig:rasta-heatmaps}
\end{figure*}

\section{Additional Results}
\label{app:additional-res}

Results for our method on the Llama-3.2 11B Vision Instruct model are provided in Table~\ref{tab:llama3.2}.

\paragraph{Discussion of results.} Unlike GPT-4, we observe a non-insignificant degradation of translation quality as measured by GEMBA and \textsc{CometKiwi}. Upon further inspection, this is because Llama-3.2-11B is overly reliant on the few-shot examples; we see many cases where the LLM includes in its translation information present in multiple few-shot examples, but not present in the original text. We use the prompt designed for GPT-4, as we want to report consistent results across LLMs. However, additional prompt tuning and a Llama-specific prompt is likely needed to prevent degradation of translation quality.

\section{Prompts}
\label{app:prompts}

The prompts for baselines and our method is included in Figure~\ref{fig:vanilla-prompt}, Figure~\ref{fig:perserve-style-prompt}, and Figure~\ref{fig:rasta-prompt}.

\paragraph{Prompt development.} To come up with the above prompts, we took the following steps:
\begin{enumerate}
    \item RASTA prompt: We had multiple rounds of iteration, where authors analyzed the results and tried to find qualitative errors (e.g. honorifics incorrectly added, no cultural knowledge incorporated, overcorrection of style, etc.) and modify the prompt accordingly. We settled on a prompt that provided the goal -- optimize for style alignment -- while still instructing the model to preserve the content.
    \item Baseline 1: Vanilla Translation Prompt: We used the most straightforward wording that an average user would be most likely to use when asking an LLM to translate text.
    \item Baseline 2: ``Preserve Style'' Prompt: We took our finalized RASTA prompt and removed the few-shot exemplars and style label components, keeping the overview and the instructions fully intact.
\end{enumerate}

\section{RASTA Examples}
Table~\ref{tab:rasta-examples} contains examples comparing RASTA translations to Baseline 1 translations. 
We include examples which show a large difference between the style of the translation from RASTA and the vanilla baseline.

For politeness, we show a Japanese sentence which is normally translated by GPT-4 to have medium politeness while Japanese native speakers say the source text was rude. Using RASTA, we can see that the translation is now significanly less polite, but still contains the same content. 

The intimacy example shows a case where the source sentence contains Spanish slang for male genitalia, which native speakers label as highly intimate. GPT-4 mistranslates this slang to its literal translation, ``whistle'', with the vanilla method. This is highlighted by the translation's low intimacy label, but using RASTA fixes this issue since the model outputs a translation with the correct intimacy, resulting in the correct slang translation.

The formality example highlights a case where a normal translation is much too formal. Native Portuguese speakers identify the source sentence as informal, and RASTA mitigates the erroneous translation by changing the phrase ``don't be under the illusion'' to ``don't kid yourself,'' an English phrase commonly used in informal settings.

\section{Human Validation Study}
\label{app:annotation}

Annotators were sourced bilingual speakers on Prolific and paid \$20 an hour. The study took, on average, 18 minutes to complete. Participants annotating for Chinese and Japanese spent 5-6 minutes longer on average than than those annotating for romance languages.

We show one example question asked in our human validation study in Figure~\ref{fig:survey-example}. We also show the instructions given to the annotators at the start of the survey in Figure~\ref{fig:survey-instructions}.

Annotator agreement, as calculated by average pairwise product-moment correlations between annotators is 0.167 for politeness and 0.247 for formality. Agreement is expectedly low due to the highly subjective nature of the task and the low number of samples. 

\begin{table}[t]
    \small
    \centering
    \begin{tabular}{llr}

    \midrule
         Style & Language & \% RASTA Favored \\
         \midrule
         & Spanish & 0.60\\
         Politeness & Chinese & 0.63\\
         & Japanese & 0.60\\

         \midrule
         & Portuguese & 0.67\\
         Formality & French & 0.63\\
         & Italian & 0.60\\
    \bottomrule
    \end{tabular}
    \caption{Language-specific results from human validation study. We have three annotators select preferred translation between RASTA and the ``preserve style'' prompting baseline. We then calculate the majority label for each of the 30 samples per language.}
    \label{tab:my_label}
\end{table}

\begin{figure*}
\centering
\begin{lstlisting}[basicstyle=\small\ttfamily]
Background
Politeness is influenced by social and cultural norms. Because of this, translating from one language to another can change politeness levels. A direct word-for-word translation does not guarantee that the politeness level of the translated text will match the politeness level of the original text.

Instructions
In this survey, you will be asked to analyze translations between English and another language and determine which translation better preserves politeness.

Important Note
Keep in mind some translations you see will contain new or modified information to reflect the politeness of the original text better.

---------------------

Please select all of the following that are TRUE about politeness. You must get this question correct to advance to the rest of the survey!

[] Politeness is a complex construct
[] Politeness is easy to translate
[] Politeness varies across cultures and languages
[] A direct word-for-word translation always preserves politeness
[] A direct word-for-word translation might change politeness
[] A translation that preserves politeness may contain new or modified information

\end{lstlisting}
    \caption{Instructions and attention check given to annotators at the start of the survey. Annotators had to pass the attention check (select options 1, 3, 5, and 6) in order to advance to the remainder of the survey.}
    \label{fig:survey-instructions}
\end{figure*}

\lstset{
    literate=
    {ñ}{{\~n}}1
    {á}{{\'a}}1
    {é}{{\'e}}1
    {í}{{\'i}}1
    {ó}{{\'o}}1
    {ú}{{\'u}}1
    {ü}{{\"u}}1
}
\begin{figure*}
    \centering
\begin{lstlisting}[basicstyle=\small\ttfamily]
Consider the following translations from English to Spanish. Please select the translation that best matches the politeness level of the original text

Original text:
I just came here for the comments. Surprisingly, there are none about hip-hop. The ridiculous amount of racist comments and articles debating whether or not hip-hop producers are musicians is beyond quantitative measurement.

Translation options:

Option A:
Simplemente vine aquí por los comentarios. Sorprendentemente, no hay ninguno sobre hip-hop. La cantidad ridícula de comentarios y artículos racistas debatiendo si los productores de hip-hop son o no músicos está más allá de cualquier medición cuantitativa.

Option B:
Solo vine aquí por los comentarios. Sorprendentemente, no hay ninguno sobre el hip-hop. La cantidad ridícula de comentarios racistas y artículos debatiendo si los productores de hip-hop son o no músicos, es más allá de cualquier medición cuantitativa.

\end{lstlisting}
    \caption{Example of survey question from our human evaluation. Option A in this example corresponds to RASTA, but we randomly shuffle the example order so that there is no pattern to RASTA being Option A or B.}
    \label{fig:survey-example}
\end{figure*}

\lstset{literate={}}

\clearpage
\begin{figure*}
    \centering
\begin{lstlisting}[basicstyle=\small\ttfamily]
Translate the following text from <Source> to <Target>.
Text: <Sample>
Output only the translation.
\end{lstlisting}
    \caption{Vanilla translation prompt used for LLMs.}
    \label{fig:vanilla-prompt}
\end{figure*}

\begin{figure*}
    \centering
\begin{lstlisting}[basicstyle=\small\ttfamily]
Your task is to translate a given piece of text from <Source> to <Target>.
When translating, you must ensure the <Style> level of the translation matches the <Style> level of the original text.
Keep in mind that <Style> varies across cultures, so a direct word-for-word translation may not always ensure the <Style> level will match that of the original text.

This is the text you need to translate:
<Sample>

Now, translate the text so that the translation also has the same <Style> level. Output only the translation.
\end{lstlisting}
    \caption{``Preserve style'' translation prompt used for LLMs.}
    \label{fig:perserve-style-prompt}
\end{figure*}

\begin{figure*}
    \centering
\begin{lstlisting}[basicstyle=\small\ttfamily]
Your task is to translate a given piece of text from <Source> to <Target>.
When translating, you must ensure the <Style> level of the translation matches the <Style> level of the original text.
Keep in mind that <Style> varies across cultures, so a direct word-for-word translation may not always ensure the <Style> level will match that of the original text.

This is the text you need to translate:
<Sample>

This text has a <Style> level of {} out of 1 in {}.

To help you translate the text in a way that preserves <Style>, here are some examples of text that have the same <Style> level in {}.
Pay attention to the way <Style> is expressed in these examples, and try to reflect it similarly in your translation.
<example 1>

<example 2>

<example 3>

<example 4>

<example 5>

Now, translate the above text to preserve the content of the message, while also making sure the <Style> level is similar to the above examples.
Specifically, the translation should also have a <Style> level of {} out of 1 in {}. Output only the translation.
\end{lstlisting}
    \caption{RASTA translation prompt used for LLMs.}
    \label{fig:rasta-prompt}
\end{figure*}

\begin{table*}[t]
\small
\renewcommand{\arraystretch}{1.2}
    \centering
    
\begin{tabular}{@{}lllrrrrrrrrrrrrr@{}}
        \toprule
        \multirow{3}{*}{\textbf{Style}} & \multirow{3}{*}{\textbf{Target}} & \multicolumn{3}{c}{\textbf{Baseline 1: Vanilla}} & \multicolumn{3}{c}{\textbf{Baseline 2: ``Preserve}} & \multicolumn{3}{c}{\multirow{2}{*}{\textbf{RASTA}}} \\
        
        & & \multicolumn{3}{c}{\textbf{Translation}} & \multicolumn{3}{c}{\textbf{Style'' Prompting}} & & \\
        
        \cmidrule(lr){3-5} \cmidrule(lr){6-8} \cmidrule{9-11}
        & & $\alignmentmetric$$\uparrow$ & CK$\uparrow$ & G$\uparrow$ & $\alignmentmetric$$\uparrow$ & CK$\uparrow$ & G$\uparrow$ & $\alignmentmetric$$\uparrow$ & CK$\uparrow$ & G$\uparrow$\\
        \midrule

\multirow{4}{*}{Politeness} & En & 0.51 & 0.77 & 90.00 & 0.53 & 0.76 & 88.60 & 0.63 & 0.72 & 82.87 \\
& Es & 0.61 & 0.72 & 91.61 & 0.61 & 0.72 & 91.91 & 0.64 & 0.69 & 89.20 \\
& Ja & 0.38 & 0.74 & 88.20 & 0.43 & 0.74 & 88.88 & 0.55 & 0.70 & 82.92 \\
& Zh & 0.56 & 0.73 & 90.52 & 0.55 & 0.72 & 90.49 & 0.70 & 0.67 & 85.32 \\
\cmidrule{2-11}
 & Avg. & 0.52 & 0.74 & 90.08 & 0.53 & 0.74 & 89.97 & 0.63 & 0.70 & 85.08\\
& RASTA $\Delta$  & \color{Green}{+ 21.2\%} & \color{Red}{-8.2\%} & \color{Red}{-5.6\%} & \color{Green}{+18.9\%} & \color{Red}{-8.2\%} & \color{Red}{-5.4\%} & -- & -- & -- \\ \midrule

\multirow{6}{*}{Intimacy} & Zh & 0.34 & 0.66 & 84.02 & 0.36 & 0.62 & 80.02 & 0.39 & 0.52 & 59.10 \\
& En & 0.50 & 0.71 & 85.34 & 0.46 & 0.67 & 79.92 & 0.51 & 0.59 & 63.70 \\
& Fr & 0.55 & 0.68 & 85.14 & 0.51 & 0.64 & 80.58 & 0.60 & 0.57 & 64.97 \\
& It & 0.55 & 0.68 & 86.87 & 0.49 & 0.64 & 81.75 & 0.57 & 0.57 & 63.37 \\
& Pt & 0.42 & 0.67 & 88.19 & 0.42 & 0.64 & 84.41 & 0.50 & 0.55 & 61.98 \\
& Es & 0.53 & 0.68 & 87.29 & 0.46 & 0.64 & 82.00 & 0.56 & 0.57 & 63.21 \\
\cmidrule{2-11}
 & Avg. & 0.48 & 0.68 & 86.14 & 0.45 & 0.64 & 81.45 & 0.52 & 0.56 & 62.72 \\
& RASTA $\Delta$  & \color{Green}{+ 8.3\%} & \color{Red}{-17.6\%} & \color{Red}{-27.2\%} & \color{Green}{+15.6\%} & \color{Red}{-12.5\%} & \color{Red}{-23.0\%} & -- & -- & -- \\ \midrule

\multirow{4}{*}{Formality} & En & 0.43 & 0.80 & 92.77 & 0.36 & 0.79 & 90.14 & 0.70 & 0.68 & 70.01 \\
& Fr & 0.42 & 0.78 & 91.74 & 0.33 & 0.77 & 90.09 & 0.60 & 0.68 & 74.85 \\
& It & 0.47 & 0.78 & 93.14 & 0.41 & 0.77 & 90.92 & 0.65 & 0.70 & 74.87 \\
& Pt & 0.46 & 0.77 & 92.94 & 0.42 & 0.77 & 91.59 & 0.62 & 0.67 & 72.84 \\
\cmidrule{2-11}
 & Avg. & 0.44 & 0.79 & 92.65 & 0.38 & 0.77 & 90.69 & 0.64 & 0.68 & 73.14 \\
& RASTA $\Delta$  & \color{Green}{+ 45.5\%} & \color{Red}{-13.9\%} & \color{Red}{-21.1\%} & \color{Green}{+68.4\%} & \color{Red}{-11.7\%} & \color{Red}{-19.4\%} & -- & -- & -- \\

        \bottomrule
    \end{tabular}
    \caption{Llama-3.2-11B: Evaluation of RASTA with prompting baselines. We measure the style alignment, $\alignmentmetric$, as well as state-of-the-art reference-free translation quality metrics GEMBA (G) \citep{kocmi-federmann-2023-large} and \textsc{CometKiwi} (CK) \citep{rei2022cometkiwi}. $\alignmentmetric$ is 1 when the interpreted style (i.e. style of translation) exactly matches the intended style (i.e. style of original text) and 0 when there is no alignment. For all metrics, higher is better.}
    \label{tab:llama3.2}
\end{table*}

\end{document}